\documentclass[10pt,journal,compsoc]{IEEEtran}

\ifCLASSOPTIONcompsoc
  \usepackage[nocompress]{cite}
\else
  \usepackage{cite}
\fi

\ifCLASSINFOpdf

\else

\fi

\usepackage{amsmath,amsfonts}
\usepackage{algorithmic}
\usepackage{algorithm}
\usepackage{array}
\usepackage[caption=false,font=normalsize,labelfont=sf,textfont=sf]{subfig}
\usepackage{textcomp}
\usepackage{stfloats}
\usepackage{url}
\usepackage{verbatim}
\usepackage{graphicx}
\usepackage{cite}
\usepackage{multirow}
\usepackage{xurl}
\usepackage{amsmath}
\usepackage{eucal}

\usepackage{amsthm}
\usepackage{amssymb}
\usepackage{bm}
\usepackage{graphicx}
\usepackage{makecell}
\usepackage{soul}
\usepackage{enumitem}
\usepackage[dvipsnames]{xcolor}
\usepackage{tcolorbox}
\tcbuselibrary{listings,breakable}

\usepackage{hyperref}
\hypersetup{
  colorlinks   = true, 
  urlcolor     = blue, 
  linkcolor    = blue, 
  citecolor    = blue 
}

\newcommand{\ie}{\textit{i}.\textit{e}.}
\newcommand{\eg}{\textit{e}.\textit{g}.}

\newcommand{\Eref}[1]{Eq.~(\ref{#1})}
\newcommand{\Fref}[1]{Figure~\ref{#1}}
\newcommand{\Sref}[1]{Section~\ref{#1}}

\usepackage{xspace}
\usepackage{pifont}
\newcommand{\name}{\textsc{FaceTracer}\xspace}

\begin{document}
\title{\name: Unveiling Source Identities from Swapped Face Images and Videos for Fraud Prevention}
\author{Zhongyi~Zhang,
Jie~Zhang,
Wenbo~Zhou,
Xinghui~Zhou, 
Qing~Guo, \\
Weiming~Zhang,
Tianwei~Zhang,
and Nenghai~Yu% 
\IEEEcompsocitemizethanks{\IEEEcompsocthanksitem Zhongyi Zhang, Wenbo Zhou, Xinghui Zhou, Weiming Zhang, and Nenghai Yu are with the School of Cyber Science and Technology, University of Science and Technology of China, Hefei, Anhui
230026, China (E-mail: ericzhang@mail.ustc.edu.cn; welbeckz@ustc.edu.cn;
zhouxinghui@mail.ustc.edu.cn; zhangwm@ustc.edu.cn; ynh@ustc.edu.cn).
\IEEEcompsocthanksitem Jie Zhang and Qing Guo are with Centre for Frontier AI Research, Agency for Science, Technology and Research (A*STAR), Singapore (E-mail: zhang\_jie@cfar.a-star.edu.sg; guo\_qing@cfar.a-star.edu.sg).
\IEEEcompsocthanksitem Tianwei Zhang is with College of Computing and Data Science at Nanyang Technological University (E-mail: tianwei.zhang@ntu.edu.sg).
\IEEEcompsocthanksitem Wenbo Zhou and Jie Zhang are the corresponding authors.
}%
}

\markboth{Journal of \LaTeX\ Class Files,~Vol.~14, No.~8, August~2015}%
{Shell \MakeLowercase{\textit{et al.}}: Bare Demo of IEEEtran.cls for Computer Society Journals}

\IEEEtitleabstractindextext{%
\begin{abstract}

Face-swapping techniques have advanced rapidly with the evolution of deep learning, leading to widespread use and growing concerns about potential misuse, especially in cases of fraud. While many efforts have focused on detecting swapped face images or videos, these methods are insufficient for tracing the malicious users behind fraudulent activities. Intrusive watermark-based approaches also fail to trace unmarked identities, limiting their practical utility. To address these challenges, we introduce \name, the first non-intrusive framework specifically designed to trace the identity of the source person from swapped face images or videos. Specifically, \name leverages a disentanglement module that effectively suppresses identity information related to the target person while isolating the identity features of the source person. This allows us to extract robust identity information that can directly link the swapped face back to the original individual, aiding in uncovering the actors behind fraudulent activities. Extensive experiments demonstrate \name's effectiveness across various face-swapping techniques, successfully identifying the source person in swapped content and enabling the tracing of malicious actors involved in fraudulent activities. Additionally, \name shows strong transferability to unseen face-swapping methods including commercial applications and robustness against transmission distortions and adaptive attacks. Our code is available at: \url{https://github.com/zzy224/FaceTracer}.

\end{abstract}

}

\maketitle

\IEEEdisplaynontitleabstractindextext

\IEEEpeerreviewmaketitle

\IEEEraisesectionheading{\section{Introduction}\label{sec:introduction}}

\IEEEPARstart{F}{ace} swapping is a well-known technique for deepfake generation, which can seamlessly transfer the identity from the target image to the source image, while the facial attributes of the source image hold intact. In recent years, popular commercial platforms like Snapchat~\cite{snap} and FaceApp~\cite{faceapp} have made face swapping accessible to millions of users worldwide, while open-source projects like DeepFaceLab~\cite{deepfacelab} and FaceSwap~\cite{faceswap} have fostered a thriving community of developers and enthusiasts. 
With the immense popularity and interest in face swapping,
this technique has been widely used in various industries such as movie production and entertainment applications. Recently, there emerge new face-swapping solutions that can even support seamless face-swapping in real time~\cite{deepfacelive}.

However, along with the commercial value and practical applications of face-swapping technology comes a significant risk. Recently, face-swapping techniques have been widely exploited in financial scams. In such cases, verifying the other party's identity in person is often difficult, leading many to rely on video calls for confirmation. Fraudsters, however, have begun using face-swapping technology to impersonate victims' family members or superiors during these calls, exploiting the trust they build to carry out scams, such as pressuring victims to transfer money~\cite{foxnews} or luring them into fake investment schemes~\cite{abc}.
In January 2024, according to Hong Kong police, a finance worker at a multinational firm was tricked into paying out \$25 million to fraudsters using face-swapping technology to pose as the company’s chief financial officer in a video conference call~\cite{cnn}.
These incidents highlight the growing concern surrounding the potential misuse of face-swapping technology and the need for countermeasures to prevent such malicious activities.

\begin{figure}[t]
\centering
\includegraphics[width=0.48\textwidth]{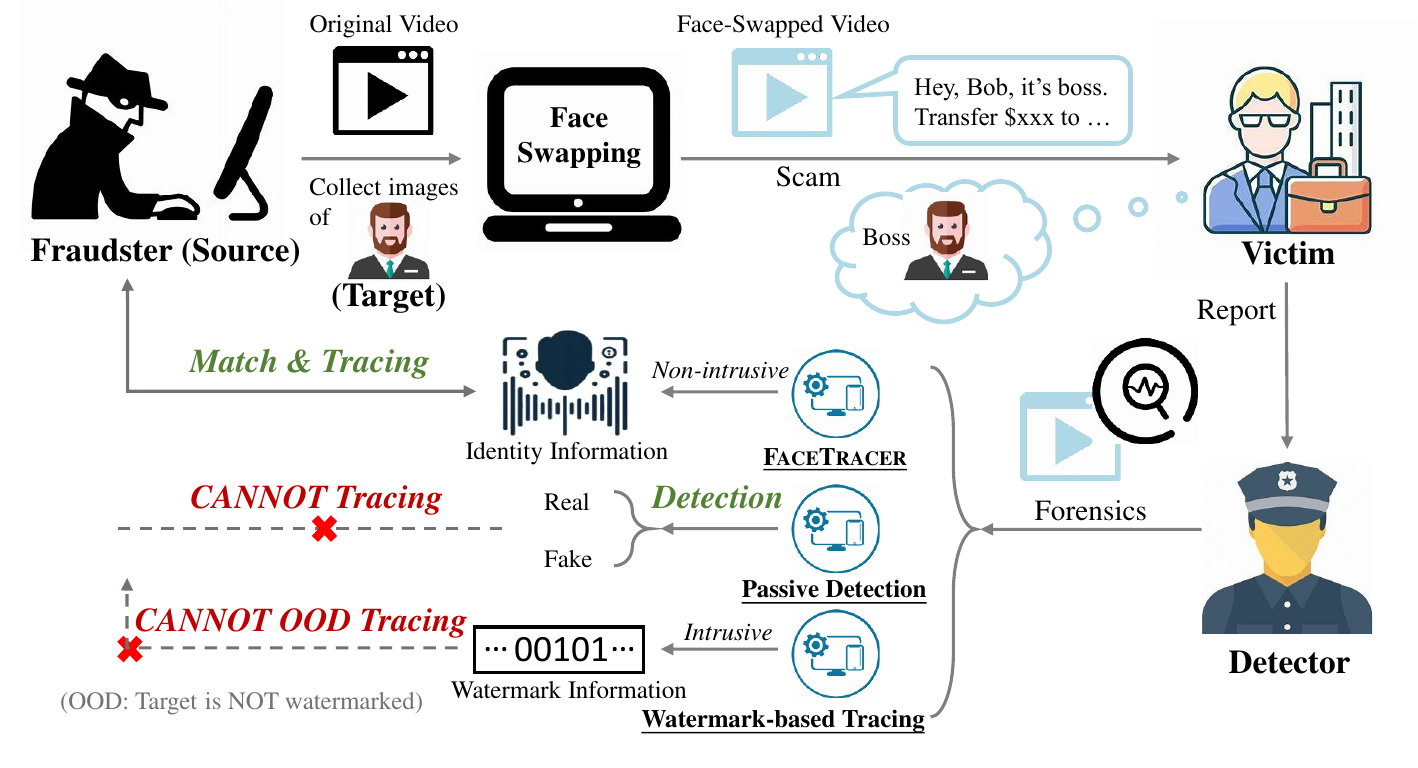}
\caption{ \name can achieve non-intrusive tracing to unveil the fraudster (source) identities for effective forensics.}
\label{fig:teaser}
\vspace{-0.5em}
\end{figure}

As illustrated in \Fref{fig:teaser}, a fraudster adopts face-swapping to replace their (source) identity with the identity of the target person (\eg, the victim’s boss) to commit fraud. To prevent such misuse and hold the fraudster legally accountable, it is crucial to uncover their true identity.
However, most existing works~\cite{sbi,caddm,altfreezing,tall} focus only on detecting whether images or videos have been face-swapped, without addressing the challenge of tracing the malicious user. Very recently, some watermark-based methods~\cite{dualdefense,faketracer} have been proposed to trace the creators of swapped face images or videos, but they need to intrusively embed watermark information into the source images or videos before the incident occurs. As a result, these approaches cannot generalize to out-of-distribution (OOD) identities (\ie, unwatermarked identities), limiting their effectiveness in real-world applications.

\textcolor{black}{While malicious attackers may download  face images from the Internet as the source identity to bypass the identity tracing, in fraud scenarios on we focus in this paper, there are two significant drawbacks to this approach: i) These images or videos can be found and used as references to verify if the content is fake. ii) These materials may contain watermarks, which can also be used to identify manipulated content. To maintain greater flexibility and control over the swapped face images or videos, attackers often need to capture new images or videos themselves to serve as the \textit{source} for face swapping.} Therefore, if we unveil the source identity from the manipulated content, we can trace the malicious fraudster.

For this, we introduces \name, the first non-intrusive tracing framework designed to effectively extract the identity information of the source person from swapped face images or videos.
As shown in \Fref{fig:teaser}, once the offense of abusing face-swapping methods occurs, \name can assist law enforcement officers in tracing the source person. To achieve it, there are two 
primary challenges: 1)
In practice, attackers may replace their original identity with various faces, resulting in swapped face images or videos visually resembling multiple individuals. In addition, the identities of the attackers can be diversified, treating this task as a classification problem by simply relabeling single swapped face images or videos is impractical. 
To address it, we opt to \ding{182} \ul{\textit{extract the identity information}} with a neural network, which is trained on a large-scale face-swapping dataset comprising over 1 million images across 30,000 identities~\cite{zhou2024rankbased}. 
2) As demonstrated in many previous studies~\cite{caddm,implicitid}, the identity information conveyed by swapped face images or videos is a hybrid of the source and target person's identities, dominated by the target person's identity. Thus, it is crucial to eliminate the influence of the target-related identity information \textcolor{black}{and capture the source-related identity information}. 
To overcome this, we design an \ding{183} \ul{\textit{identity information disentanglement module}}, targeting at retaining only the source-related identity information while removing target-related identity information. 

Empowered by the above two designs, \name can trace the attacker by comparing the similarity between the extracted source identity information and the identity information stored in \textbf{the identity pool}. Notably, in the real-life forensics process we face an open-world problem, where the information relevant to the attacker may not be presented in the training data. Even under such cases, \name is still able to extract the identity information of the attacker.

Extensive experiments are conducted 
over four popular face-swapping methods that explicitly separate identity and attribution information of faces: HiRes~\cite{HiRes}, FaceShifter~\cite{faceshifter}, SimSwap~\cite{simswap}, and InfoSwap~\cite{infoswap}. \name demonstrates its superior ability to extract identity information of the source person from swapped face images or videos under different forensic conditions, and has good transferability for identities and methods that have not appeared in the training process. In addition, evaluation on other two face-swapping methods (\ie, MegaFS~\cite{megafs} and DiffSwap~\cite{diffswap}) shows that \name also achieves good performance against face-swapping solutions that do not explicitly separate attribution and identity information. Notably, \name performs well on commercial apps such as Faceover~\cite{faceover} and DeepFaker~\cite{deepfaker}. 
We also investigated the robustness of \name against distortions that may be encountered during various network transmissions, such as JPEG, color jittering, etc.  \textcolor{black}{Saliency map visualization is also conducted to better understand the regions \name focuses on during tracing.}
Finally, we further analyze the impact of different backbones, the disentangle networks, and the scale of the identity pool, adaptive evaluation on \name is also discussed.

In a nutshell, our main contributions could be summarized as follows:

\begin{itemize}[leftmargin=*]
    \item We proposed the first non-intrusive tracing framework, \name, which can extract the source identity information from  the swapped face images or videos, enabling effective forensics. 
    
    \item \name consists of two main parts: an identity extractor trained on large-scale dataset and an effective disentangle network to maximally eliminate the influence from the target-related identity information. Based on these, \name can extract identity information that is highly related to the source person.

    \item \name holds general effectiveness among different face-swapping methods, including four that explicitly and two that implicitly disentangle identity and attribute information, and two commercial apps, even those that have not been seen during the training phase.
    
    \item \name is robust against different distortions and even adaptive attacks, making it effective in practice.
\end{itemize}

\section{Background}

\subsection{Face Swapping}
\label{sec:fs}

\begin{figure}[t]
\centering
\includegraphics[width=9cm]{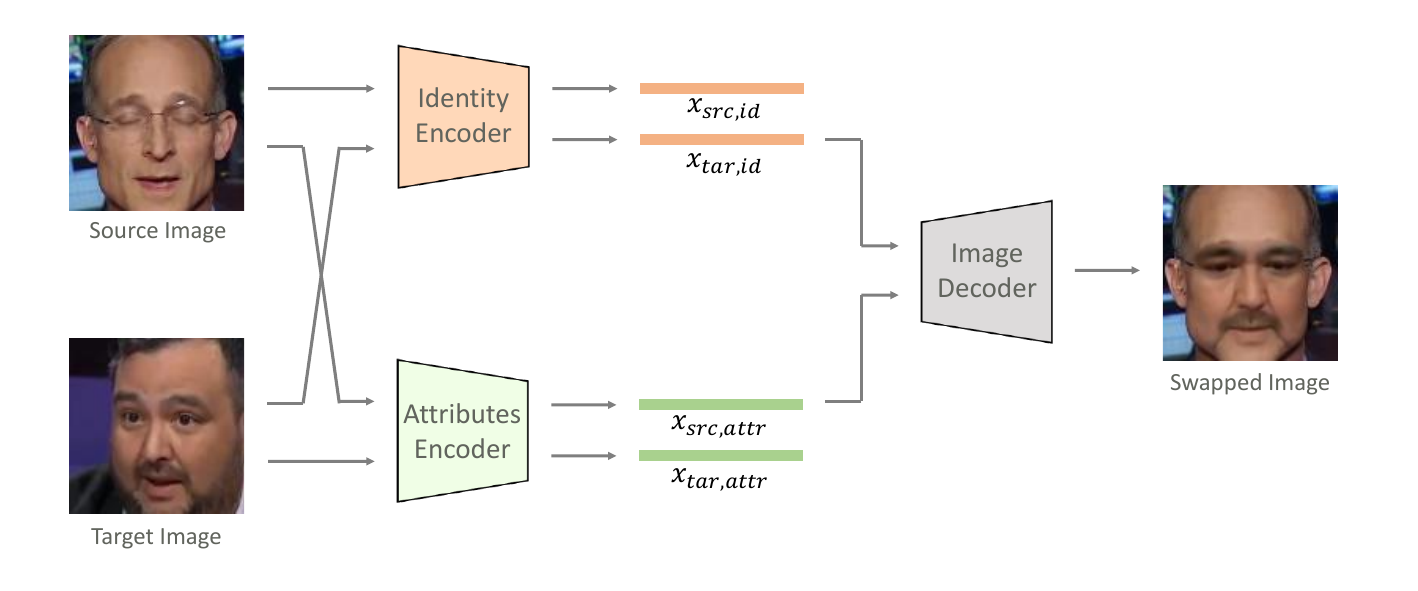}
\caption{Illustration of traditional face swapping techniques.}
\label{fig:process}
\end{figure}

In the study of the human face, 
the entirety of facial information can be categorized into two main components: identity information, determining ``who this person is", and facial attributes, encompassing expressions, hairstyles, gaze direction, etc. 
The objective of face swapping is to make the person in an image or a video resemble the target person while retaining the facial attributes of the source person unchanged. This can lead others to mistakenly believe that the person in the image or video is the target person.
Figure~\ref{fig:process} describes the process of traditional face-swapping methods in general. 
Formally, an identity encoder $\mathcal{E_{\textit{id}}}$ aims to extract the identity embedding  $x_{src,id}$ and $x_{tar,id}$  from the source image and the target image, respectively. Similarly, we can obtain the attribution embedding $x_{src,attr}$ and $x_{tar,attr}$. Then, feeding $x_{src,attr}$ and $x_{tar,id}$ into the face image decoder $\mathcal{D}$ together, the swapped image is acquired. 

It is noted that achieving the disentanglement of identity and attribute information plays a crucial role during the face swapping process. There are some methods that aim to explicitly disentangle the identity and attribute information with well-designed networks.
For example, some methods~\cite{dvp,pagan} leverage off-the-shelf 3D-based models~\cite{flame,d3dfr,deca,emoca} to extract facial attributes, which only remain the structure of human faces and neglect the corresponding texture information. Similarly, HiRes~\cite{HiRes} performs face-swapping by leveraging the generative networks and 2D attributes extraction methods. These methods separately extract identity and attribute information with off-the-shelf networks and combine them to produce swapped face results. Besides, some other methods~\cite{simswap,faceshifter,infoswap,smoothswap,hififace} first train a single conditional GAN to reconstruct face images with two branches before producing face-swapping, namely, the identity extraction branch and the attributes extraction branch. That is to say, the disentanglement of identity and attribute information is accomplished through the two branches.

Recently, as the capability of generative networks increases, some methods~\cite{megafs,diffswap,diffae} have also adopted approaches that do not explicitly disentangle identity and attributes. For example, MegaFS~\cite{megafs} assumes that some existing generative networks can inherently separate identity and attribute information, \ie, different latent space layers of StyleGAN~\cite{stylegan,stylegan2} correspond to different levels of semantics from coarse to fine. Moreover, based on the success of diffusion models~\cite{ddpm,ddim}, DiffSwap~\cite{diffswap} leverages the latent space of diffusion models as it preserves the layout of the source image. Instead of training the network to individually extract the facial attributes, these methods directly inject the target identities $x_{tar,id}$ into the generative network to generate the swapped faces.
In \Sref{sec:fsm}, we adopt 4 open-sourced face-swapping methods that explicitly disentangle identity and attribute information (\ie, HiRes~\cite{HiRes}, FaceShifter~\cite{faceshifter}, SimSwap~\cite{simswap}, and InfoSwap~\cite{infoswap}) and two open-sourced face-swapping methods that implicitly disentangle such information (\ie, MegaFS~\cite{megafs} and DiffSwap~\cite{diffswap}) for comprehensive evaluation.

\subsection{Identity Extraction}
\label{sec:extraction}

In our scenario, we need to extract the identity information from the suspected face image at the first step. In the process of extracting face identity from an input face image, the initial step involves detecting faces within the image using face detection techniques like MTCNN~\cite{mtcnn} or Face Attention Network (FAN)~\cite{fan}. Once detected, bounding boxes are used to isolate these faces. Subsequently, the faces are cropped and resized to a standardized size, typically $112\times112$ pixels. Afterwards, the cropped and scaled faces $I$ are fed to the identity extraction network. The majority of current identity extraction networks employ the ResNet~\cite{resnet} family as their backbone architecture, while some networks also utilize the Vision Transformer (ViT)~\cite{vit} as an alternative backbone. The same thing is that the identity information is usually represented as the output of the network, typically a 512-dimensional real vector. Thus, an identity information extraction network could be formulated as:
\begin{equation} \label{eq:f}
\mathcal{F}: I \rightarrow x \in \mathbb{R}^{1\times512}.
\end{equation}

While various identity information extraction networks may share similar or identical backbones, the final representation of $x$ can be different significantly due to the diverse training strategies.
Briefly, training an open-set identity extraction network can be regarded as a classification task. Intuitively, the simplest identity extraction network consists of a softmax activation with a classification layer, employing the cross-entropy loss function to constrain the representation distribution of extracted identity information, which could be formulated as:
\begin{equation}
\label{eq:softmax}
\mathcal{L}=-\frac{1}{N}\sum_{i=1}^{N}y_i\log\frac{e^{\bm{W_{y_i}}^Tx_{i}}}{\sum_{j=1}^{N}e^{\bm{W_j}^Tx_{i}}},
\end{equation}
where $x_i$ denotes the real identity vector extracted from the network, and $\bm{W_i}$ denotes the linear mapping layer that converts the identity vector to a prediction of the likelihood that the current identity vector belongs to the label $i$. However, advancements in this field have introduced several techniques to enhance performance by modifying the loss function and decision boundaries. For instance, SphereFace\cite{sphereface} introduces a multiplicative angular margin, CosFace\cite{cosface} introduces an additive cosine margin, ArcFace\cite{arcface} introduces an additive angular margin, and AdaFace\cite{adaface} introduces adaptive angular margin. These methods aim to make the extracted identity features more compact in the angular feature space, ultimately improving the performance of identity extraction.

\subsection{Feasibility of Extracting the Source Identity}
\label{sec:feasibility}

\begin{figure}
\centering
\includegraphics[width=0.4\textwidth]{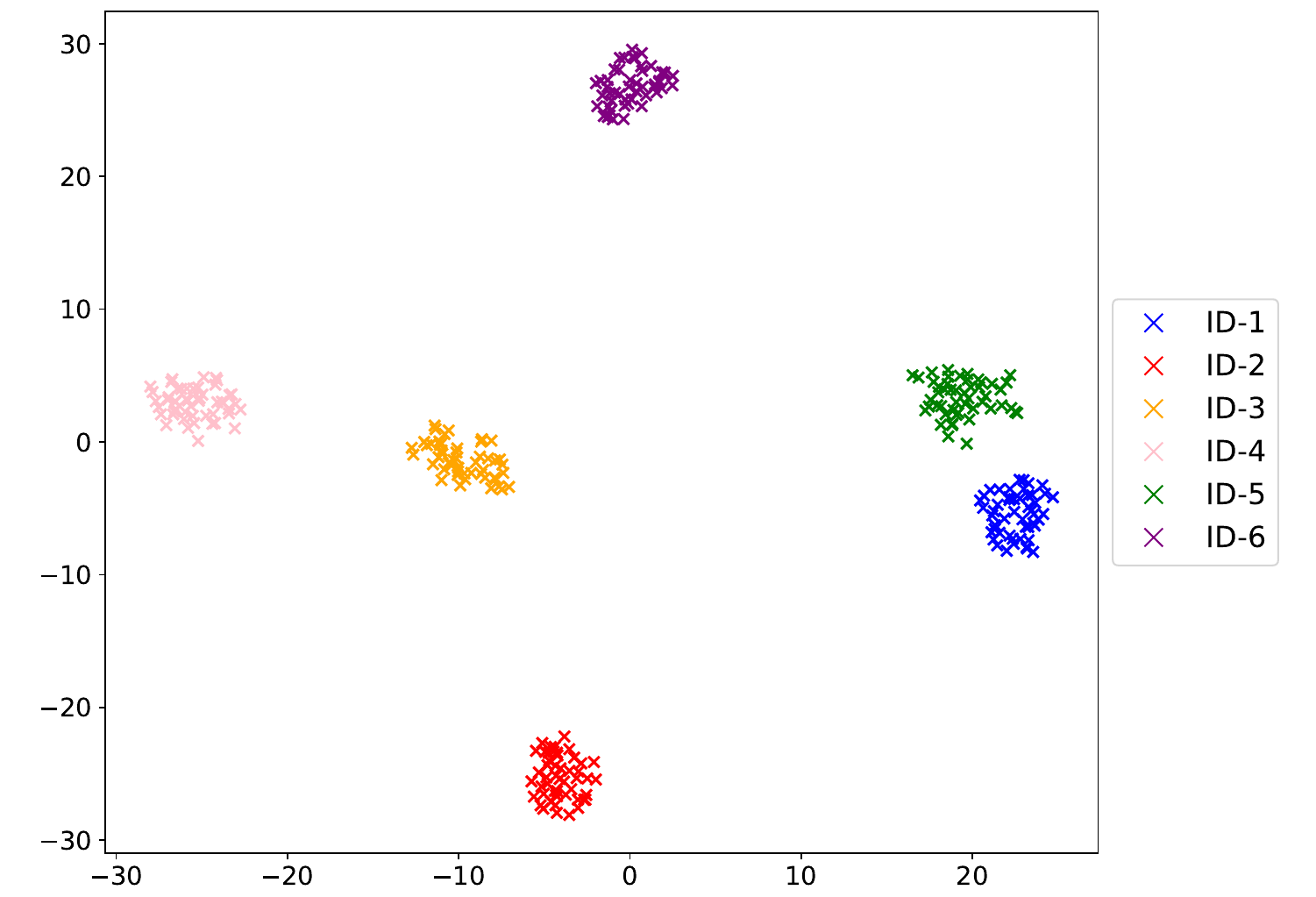}
\caption{T-SNE visualization of the attribution embedding extracted by EMOCA~\cite{emoca} from the generated images (the same attributions but different identities). The distinguishable cluster verifies that some identity information is preserved in the extracted attribution embedding. }
\label{fig:tsne-exp}
\vspace{-0.5em}
\end{figure}

\name aims to extract the source identities $x_{src,id}$ from the swapped face images or videos, and the feasibility of this goal can be attributed to two main reasons. 

First, current methods are unable to perfectly disentangle identity and attribute information, which means that the extracted source attributes $x_{src,attr}$ contains some source identity information. Therefore, when performing face-swapping operation, part of the source identity information will be inevitably retained.
Below, we adopt the state-of-the-art attributes extraction network EMOCA~\cite{emoca} and conduct experiments to verify that the extracted attribute information inherently contains some identity information. Here, we consider a special case, wherein all faces share the very similar attributions but different identity information. In consideration of the difficulty of collecting data with the same attributes in reality, we utilize StyleGAN~\cite{stylegan,stylegan2} to demonstrate this. 
It is known that different latent layers in StyleGAN control different levels of the semantic  information~\cite{tedigan,styleclip}. Thus, we fixed the first 12 layers of the latent codes in a standard 18-layer StyleGAN network to produce images with similar attributes but different identities. 
For more images, we fix the latent codes and adjust the noise injected to StyleGAN, and visual examples of the generated images are provided in the supplementary material.
Finally, we use EMOCA~\cite{emoca} to extract the attributes from the above generated images, and find that the expression cosine similarity between different identities is greater than 0.9. Thereafter, we performed t-SNE analysis on the expression embedding extracted from 6 different identities, within each identity 50 images were generated by adjusting the noise input. The analysis result of expression embedding is presented in Figure~\ref{fig:tsne-exp}, where a clear clustering exists between the different identities with highly similar attribute. In addition, we trained an SVM to classify these data and obtained over 99\% classification accuracy.
All the results consistently verify that the extracted attribute information contains some identity information.

Second, another important step in the face-swapping is the extraction of the target identity information $x_{tar,id}$. However, this process is also imperfect. To illustrate this, we randomly selected several videos with single faces (\ie, the same identity) and utilized the widely-used identity extraction network, ArcFace~\cite{arcface}, to extract the identity information. We then calculated the cosine identity similarity between each frame with different poses and expressions, obtaining a cosine identity similarity of approximately 0.8. This observation highlights the imperfect nature of identity information extraction. Due to this imperfection, the identity in the swapped face result cannot be seamlessly replaced with the identity of the target person. In general, the identity in the swapped face result will behave as a hybrid identity that is a mixture of most of the target identity and a small portion of the source identity. 

\subsection{Face Identification}

Face identification systems are vital tools for the tracing task, whose objective is to identify the most similar identity from an input face image $I$ to one of the identities within an existing  identity pool holding $N$ identities, \ie, $\{x_n\}_{n\in N}$. This process enables the determination of the specific person in the pool corresponding to the input face image. Typically, we use the cosine similarity metric $cos(\cdot,\cdot)$ to measure the similarity of different identity information, so face identification system can be formalized as follows:
\begin{equation}
\mathop{\arg\max}\limits_n cos(\mathcal{F}(I), x_n).
\end{equation}
However, face-swapping techniques can mislead face identification systems, making it believe that the identity in the image belongs to someone else, leading to malicious events such as fraudulent and scapegoating.
In this paper, \name first extract the source identity information from the suspected face, and then compare the it with the one in the identify pool for tracing the malicious user.

\section{Threat Model}

In this section, we first give the problem formulation, and then clarify the ability and goals of both \textit{Attacker} and \textit{Defender}. Without loss of generality, we refer to the party who uses face-swapping methods to convert their identity to the target identity as the \textit{Attacker} and the party who wants to extract the source identity as the \textit{Defender}. Besides, three practical scenarios are also illustrated for subsequent evaluation.

\subsection{Problem Formulation}
\label{sec:formulation}

Most contemporary face-swapping methods primarily operate at the image level, although some extend to video processing, these methods typically swap faces in each frames individually before assembling them into a cohesive video sequence. Consequently, our discussion predominantly centers on swapped face images. Nevertheless, in \Sref{sec:video}, we delve into the impact of varying input frame counts when extending \name to video level.

\begin{figure}[t]
\centering
\includegraphics[width=8.5cm]{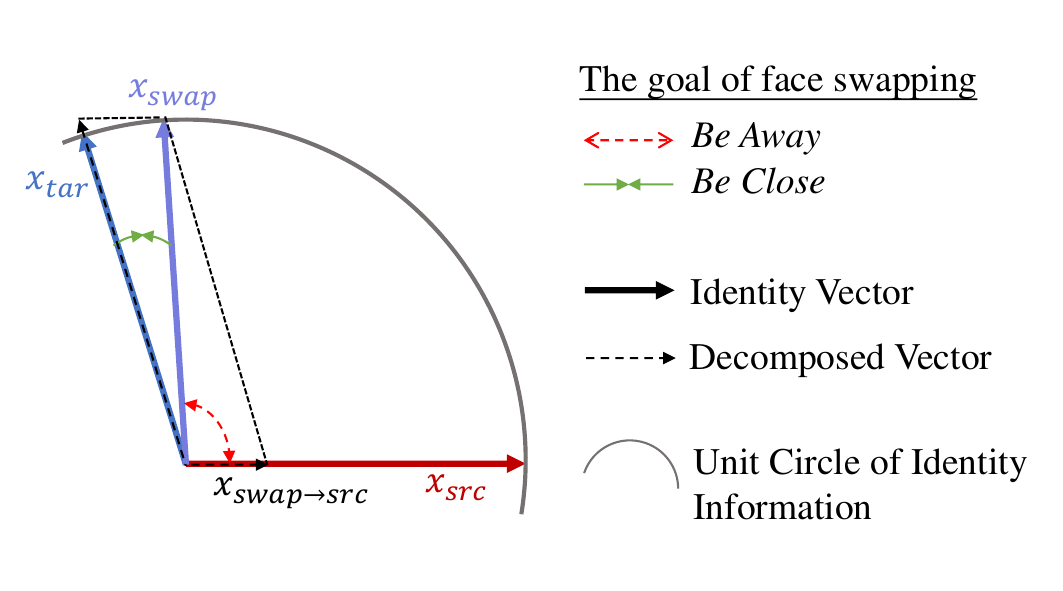}
\caption{Relationships between identity information during face-swapping. All identity information is normalized and located on the unit circle. The identity information of the swapped face image is dominated by the target identity information and still contains some source identity information.}
\label{fig:disentangle}
\end{figure}

Given a raw facial image $I_{src}$ of the source person, the swapped face image $I_{swap}$ was generated through face-swapping method to mimic the identity of the target person $x_{tar,id}$, \ie,
\begin{equation}
I_{swap}=\mathcal{S}(I_{src},I_{tar}),
\end{equation}
where $\mathcal{S}(\cdot,\cdot)$ denotes arbitrary face-swapping methods. 
As shown in \Fref{fig:disentangle}, face-swapping methods aim to significantly increase the identity similarity between $I_{swap}$ and $I_{tar}$, \ie, 
\begin{equation}
\textcolor{black}{cos(\mathcal{F}(I_{swap}),\mathcal{F}(I_{tar}))\rightarrow 1},
\end{equation}
while pursuit the decrease of the identity similarity between $I_{swap}$ and $I_{src}$, \ie, 
\begin{equation}
\textcolor{black}{cos(\mathcal{F}(I_{swap}),\mathcal{F}(I_{src}))\rightarrow 0},
\end{equation}
making it difficult for the defender to unveil the identity of the attacker. As we have discussed in \Sref{sec:feasibility}, due to the imperfectness of the face-swapping process, the swapped face result will form a hybrid identity that is a mixture of both the source and target identities.

In this paper, \name is intended to extract the source identity information from the suspect face image $I$, \ie, 
\begin{equation} \label{eq:e}
\mathcal{E}: I\rightarrow x_{src,id} \in \mathbb{R}^{1\times512}.   
\end{equation}

When the fed image is an un-swapped face, the goal of \Eref{eq:e} is the same with \Eref{eq:f}. If the suspected face image is a swapped one, $\mathcal{E}$ focuses on extracting identity of the source person (\ie, the malicious user), while $\mathcal{F}$ targets extracting the general identity of the swapped image $I_{swap}$ itself. In the following part, we adopt the straightforward solution $\mathcal{F}$ as the baseline for comparison.

\subsection{Attacker's Ability and Goal }

We assume that the attacker possesses facial images $\{I_{tar}^a\}$ of the target person, and can extract target person's identity information $x_{tar,id}$ from them. The attacker then manipulates carefully crafted images $I_{{src}}$ or videos $V_{{src}}$, employing face-swapping methods to replace their faces within these media with the face of the target person. This process aims to create swapped face results $I_{swap}$ or $V_{swap}$. 
It is worth noting that, due to the design that explicit disentangle identity and attribute information used in most of the current face-swapping methods, we assume that the model used by the attacker adopts this design. 
Besides, we also evaluate our transferability to face-swapping methods that do not explicitly disentangle identity and attribute information, and even commercial apps in Section~\ref{sec:gen}.

\subsection{Defender's Ability and Goal}
\label{sec:defender}

We assume the defender acquires an image $I$ or video $V$, recognizing it as potentially face-swapped but unaware of the face-swapping method used in $I$ or $V$. The defender may also obtain some reference face images of the target person $\{I_{tar}^r\}$,
which allows the extraction of the target person's identity information $\hat{x}_{tar,id}$. Notably, these acquired face images do not need to be those held by the attacker (\ie, $ \{I_{tar}^r\} \neq \{I_{tar}^a\} $) and $\hat{x}_{tar,id}$ need not be exactly the same as $x_{tar,id}$. 

The defender's objective is to train a \name model that could extract the identity information of the source person from the input image. 
After completing the training phase of \name, the defender can obtain an identity pool for identity matching, which is practical in reality,  
\eg, the Police Force could use all the face images in their database to create such an identity pool. To preserve privacy, only identity embeddings rather than face images are stored in the identity pool. Subsequently, the defenders could obtain the identification network to match extracted identity information with those in the identity pool, which offers the possibility to trace the attacker. 
\textcolor{black}{We posit that \name, at its core, functions as a specialized form of face recognition. Therefore, privacy-preserving methods such as adding adversarial noise~\cite{advface} can be seamlessly integrated into our training process. This integration would enhance \name's privacy protection capabilities without compromising its effectiveness in detecting face-swapping fraud.}

For forensics integrity, the defender shall extract the correct identity information from un-swapped faces, which is identical to the true identity information of these images or videos.

\begin{figure}[t]
\centering
\includegraphics[width=8.5cm]{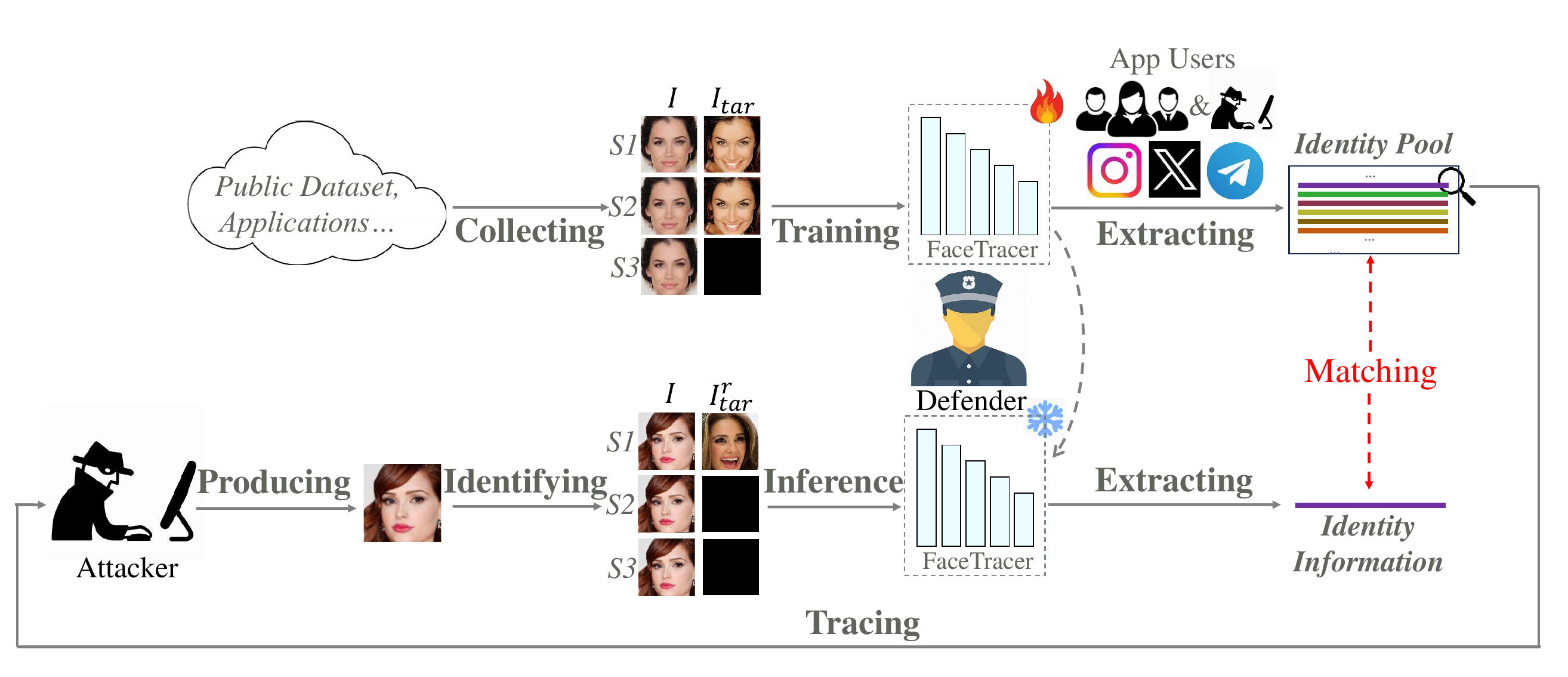}
\caption{Illustration of three scenarios of \name. 
Different Scenarios differ in the access to the reference images $I_{r}$ during the training and testing phases. The reference images $I_{r}$ denote face images of the target person used as reference.
The blank images are used in the absence of reference images. }
\label{fig:scenario}
\end{figure}

\subsection{Practical Scenarios} \label{sec:sce}

In practice, the defender may have access to some reference images of the target person to support forensics (\ie, $\{I_{tar}^r\}$), and \Eref{eq:e} can be specified as follows:
\begin{equation}
\max cos(\mathcal{E}(I,*|\theta), x_{src,id}),
\end{equation}
where $\theta$ is the parameters of the \name model and $*$ denotes the options of the reference images. Based on the defender's access to reference images during the training and inference phases, we classify \name into three main scenarios: full-reference scenario, half-reference scenario and none-reference scenario, as shown in Figure~\ref{fig:scenario}. 
\begin{tcolorbox}[colback=gray!25!white, size=title, breakable, boxsep=1mm, colframe=white, before={\vskip1mm}, after={\vskip0mm}]
\textbf{Note}: In all three scenarios, there is no identity overlap between the training and inference phase. That is to say, \name learns to disentangle source and target identity rather than memorizing some identities.
\end{tcolorbox}

\noindent\ul{\textbf{Full-reference Scenario (S1)}}: In full-reference scenario, the target person's face images $\{I_{tar}\}$ used for creating swapped face images or videos are available during the training phase, and reference images of the target person $\{I_{tar}^r\}$ are also available during the inference phase. Furthermore, to ensure that \name can correctly extract identity information from the images without face-swapping, some raw images without face-swapping are used during the training phase, where the corresponding reference images are replaced by the blank images of the same size.
In this scenario, defenders like law enforcement agencies have identified the specific target identities used in the face-swapping process. This situation typically arises during retrospective investigations. 

\textit{Case Example:} 
Consider a case where an attacker creates a fraudulent video by swapping the face of the victim's boss. Once the victim reports the fraud to authorities, the defender can easily determine the target person and gather reference images. This knowledge of the target identity significantly aids in the investigation and analysis of the face-swapped content.

\noindent\ul{\textbf{Half-reference Scenario (S2)}}: The main difference between S2 and S1 is that the reference images of the target person $\{I_{tar}^r\}$ are unavailable during the inference phase. The training phase of S2 is identical to that of S1. During inference phase, the reference images $\{I_{tar}^r\}$ are replaced with blank images of the same size.
\textcolor{black}{In this scenario, the defender is not determined who exactly the victim is and therefore uses a blank image as a substitute for the target face image in the inference phase. As \name learns to extract source-relevant identity information in the swapped face images, the reference images in the inference phase are helpful but not essential.}

\noindent\ul{\textbf{None-reference Scenario (S3)}}: In none-reference scenario, not only $\{I_{tar}^r\}$ are unavailable during the inference phase, but $\{I_{tar}\}$ are also unavailable during the training phase, wherein reference images are replaced with blank images of the same size in both the training and inference phases.
While \name obtain less information during the training phase, Table~\ref{tab:main} shows that reference images are not necessary to the training phase. In other words, \name can disentangle the source and target identity well even without the reference information of the target identity.

\begin{figure*}[t]
\centering
\includegraphics[width=17cm]{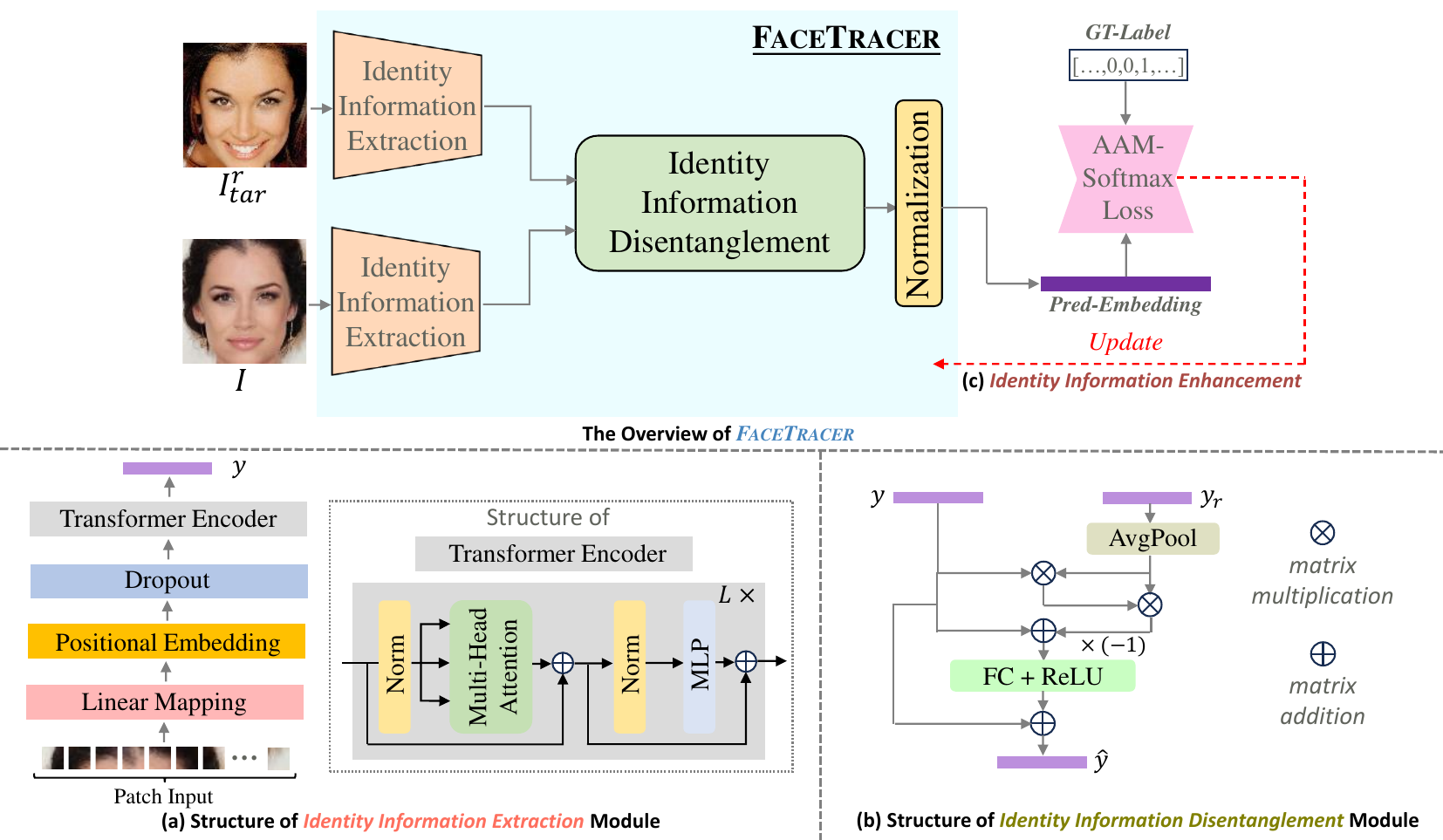}
\vspace{-1em}
\caption{\name consists of three designs: a) \textit{identity information extraction} module aims to extracts crude identity embedding from the input image, b) \textit{identity information disentanglement} module eliminates the influence of the identity information from the target person, and c) \textit{identity information enhancement} module further enhances the discriminative capability. \textcolor{black}{The red dotted line indicates the optimization of \name with the identity information enhancement module, which is also known as the AAMSoftmax loss in Eq.~(\ref{eq:aamsoftmax}).}}
\label{fig:structure}
\vspace{-0.5em}
\end{figure*}

\section{\name}

The purpose of \name is to design an identity information extractor of the source person for any input image. To achieve this, we referred to previous work~\cite{Phan_2024_WACV} on model designs for extracting identity information and further designed a disentangle module to extract the identity information of the source person rather than hybrid identity information. We also utilized a normalization layer followed by an additive angular margin softmax activation to enhance the performance of \name. Noted that the normalized output will be treated as the extracted identity information of the source person. The training phase of \name consists of three designs: 

\begin{itemize}[leftmargin=*]
    \item \textit{\textbf{\textcolor[RGB]{0,0,0} {Identity Information Extraction}}} module $\mathcal{E}_1(\cdot)$  extracts the crude identity information from the input image.
    
    \item \textit{\textbf{\textcolor[RGB]{0,0,0}{Identity Information Disentanglement}}} module  $\mathcal{E}_2(\cdot)$ is fed with the extracted identity information. Then, it cleanse the relevant portion of the target person's identity information from the hybrid identity information and retain the identity information of the source person.

    \item \textit{\textbf{\textcolor[RGB]{0,0,0}{Identity Information Enhancement}}} is a customized loss $\mathcal{L}$ that further reduces the distance between the extracted identity information and the ground-truth identity information, and increases the distance between the generated identity information and other identity information, thus enhancing the discriminative ability of facial identification systems.
\end{itemize}
In a nutshell, \name can be formulated as $\mathcal{E}(I, *|\theta) = \mathcal{E}_2(\mathcal{E}_1(I), *)$, trained with loss function $\mathcal{L}$. Figure~\ref{fig:structure} showcases the designs of \name, and we will describe them in details below.

\subsection{Identity Information Extraction}
\label{sec:IIE}

We leverage the ViT-S model from the vision transformer (ViT) family~\cite{vit} as the default backbone for identity information extraction. As shown in Figure~\ref{fig:structure}~(a), due to the patch-based input structure of the Transformer, the input image is first divided into $s\times s$ patches. After obtaining the patches, they are mapped to $d$-dimensional token vectors through a linear mapping layer, resulting in a total of $n$ $d$-dimensional tokens, where $n$ is calculated from $n=\lfloor \frac{h*w}{s*s} \rfloor$. Subsequently, positional embedding is applied to these patches followed by a dropout layer with the dropout rate $p$ to enhance the model's performance. ViT-S comprises $l$ transformer blocks in total, each consisting of two parts: (i) a multi-head self-attention layer with normalized input and the skip-connection mechanism and (ii) a 2-layer MLP with normalized input and the skip-connection mechanism. The input and output of each Transformer block are the same, and the self-attention mechanism offers the capability to extract identity information from the image. Generally, the identity information extraction block converts the input image into an $n\times d$-dimensional identity information vector. We formulate the identity information extraction module as:
\begin{equation}
\bm{y}=\mathcal{E}_1(I)\quad \bm{y}\in\mathbb{R}^{n*d},\forall I. 
\end{equation}
The hyper-parameters $s$, $d$, $h$, $w$, $p$, and $l$ are set to $9$, $512$, $112$, $112$, $0.1$, and $12$ by default in our framework.

As discussed in \Sref{sec:extraction}, utilizing ResNet~\cite{resnet} for identity information extraction remains a viable approach. While the ViT family has demonstrated superior performance in various tasks, many identity information extraction networks still opt for the ResNet as the backbone, exemplified by ArcFace\cite{arcface}. Unlike the ViT series, ResNet directly processes the image as input to the network, extracting high-level semantic information through several residual blocks. In Section~\ref{sec:ablation}, we will explore the implications of replacing the backbone with ResNet18 on the model's performance.

\subsection{Identity Information Disentanglement}
\label{sec:IID}

The identity information disentanglement module is the key to extract the identity information of the source person from the input image. As mentioned in \Sref{sec:formulation} and Figure~\ref{fig:disentangle}, the swapped face image has identity information that is similar to the identity information of the target person but is distant from that of the source person. Therefore, simply utilizing an identity extraction network $\mathcal{E}_1$ to extract the identity information of input images does not effectively obtain the identity information of the source person. To address this issue, we aim to explicitly eliminate the influence of the target person's identity information by designing a dual-input network. This approach maximizes the relevance between the extracted identity information and the identity information of the source person. Inspired by REVELIO~\cite{revelio}, as shown in Figure~\ref{fig:structure}~(b), we designed a disentanglement module to explicitly disentangle identity information. Under this design, the inputs of this module are $\bm{y}$ and $\bm{y}_{r}$, calculated from the following equations:
\begin{equation}
\bm{y}=\mathcal{E}_1(I),\quad \bm{y}_{r}=\mathcal{E}_1(I_{tar}^r),
\end{equation}
where $I$ denotes the suspected input image and $I_{tar}^r$ denotes the reference image \textcolor{black}{of the target person}, noted that $I_{tar}^r$ could be blank image in scenarios S2 and S3. Because the original identity information vector $\bm{y}$ is added in a residual learning fashion,  the total identity information disentanglement module $\mathcal{E}_2$ could be formulated as:
\begin{equation}
\hat{\bm{y}}=\mathcal{E}_2(\mathcal{E}_1(I),*).
\end{equation}
In addition, in distinction to explicitly disentangle the identity information of the target person from the extracted hybrid identity information, it is also viable to use some implicit disentanglement methods such as cross-attention. We will discuss the impact of this implicit disentangle approach on the model performance in Section~\ref{sec:ablation}.

\subsection{Identity Information Enhancement}

As mentioned in Section~\ref{sec:formulation}, conventional identity extraction networks typically employ a normalized softmax activation with a cross-entropy loss function. However, relying solely on this setting may not adequately differentiate between extracted identity information, potentially impacting the identification model's performance. To address this, we took inspiration from ArcFace~\cite{arcface} and implemented an additive angular margin softmax (AAM-Softmax) activation. This approach aims to better separate the extracted identity information of different source person on the unit circle. The loss function of \name is formulated as:
\begin{equation}
\label{eq:aamsoftmax}
\mathcal{L}=-\frac{1}{N}\sum_{i=1}^{N}y_i\log\frac{e^{s(cos\phi_{y_i}+m)}}{e^{s(cos\phi_{y_i}+m)}+\sum_{j=1,j\neq y_i}^{N}e^{scos\phi_j}},
\end{equation}
where $x_i=\mathcal{E}(I_i,*|\theta)$ is the 512-dimensional real identity vector output of the previous network and $cos\phi_j$ is calculated from:
\begin{equation}
cos\phi_j=\frac{\bm{W}_j^Tx_i}{||\bm{W}_j||*||x_i||}\quad \forall j,\forall x_i. 
\end{equation}
Here $s$ and $m$ are parameters that control the distribution of extracted identity information vectors, we set $s$ and $m$ to $64$ and $0.5$ by default. Additionally, while the backbones of the extraction model may vary in structure, they all produce outputs of the same size: 512-dimensional real vectors. Therefore, we apply the AAM-Softmax activation as identity information enhancement to the outputs of every architecture and the formulation of the loss function will stay the same.

\subsection{Extension to Swapped Face Videos}
\label{sec:ext}

Although \name mainly functions at the image level, it can easily be extended to videos. For a suspect video $V$, we can extract source IDs from each frame or selected frames. This allows the inference process to revert to image-level extraction, where reference images ${I_{tar}^r}$ or blank images can also assist in forensic analysis. Further details can be found in \Sref{sec:video}.

\section{Experiment}

\subsection{Experimental Setting}

\noindent\textbf{Face-swapping Methods.}
\label{sec:fsm}
We selected four widely-used face-swapping methods that explicitly disentangle identity and attribute information to separately train our \name, and demonstrate transferability among each other.

\begin{itemize}[leftmargin=*]
    \item \textbf{HiRes}~\cite{HiRes}
    transfers different levels of attributes by three modules and learns a structure transfer direction in the latent space of StyleGAN~\cite{stylegan,stylegan2}. The face-swapping result is produced by swapping the identity-relevant latent codes of the target image and the refined latent codes corresponding to face attributes of the source image.
    
    \item \textbf{FaceShifter}~\cite{faceshifter} is an occlusion-aware face-swapping method, which first uses an adaptive embedding integration network (AEI-Net) to generate the first-stage swapped face image, followed by a heuristic error acknowledging refinement network (HEAR-Net) to produce better face-swapping result with occlusions.
    
    \item \textbf{SimSwap}~\cite{simswap} is based on an Encoder-Decoder architecture. It first utilize an encoder to extract attributes feature from the source image and uses the ID Injection Module (IIM) to transfer the identity information from the target face into the extracted attributes feature. After that, the decoder restores the modified features to the result swapped face image.
    
    \item \textbf{InfoSwap}~\cite{infoswap} aims to disentangle identity-related information from facial images and then swap this information between different images while preserving other attributes. It also employs an encoder-decoder architecture as SimSwap to reconstruct images from the feature. Besides, InfoSwap applies an information bottleneck to the architecture, forcing the encoder to learn a compact representation that retains only the most relevant information for identity swapping and disentangle identity-related features from other attributes.
\end{itemize}

To further assess the transferability of \name on methods that do not explicitly decouple identity and attribute information, we also tested the model using swapped face images from the following two methods:
\begin{itemize}[leftmargin=*]
    \item \textbf{MegaFS}~\cite{megafs} produces face-swapping in the latent space of StyleGAN ~\cite{stylegan,stylegan2} without extracting any identity or attribute information. The swapped face result was directly generated by the StyleGAN2 generator with the manipulated latent codes. 
    
    \item \textbf{DiffSwap}~\cite{diffswap} is a face-swapping method based on diffusion models~\cite{ddpm,ddim}, which first extract identity information from the target image and inject the identity information into the conditional reverse diffusion process of the source face. Noted that the attribute information is not explicitly extracted during this process.
\end{itemize}

Moreover, we evaluated \name on two commercial face swapping apps, \ie, Faceover~\cite{faceover} and DeepFaker~\cite{deepfaker}.

\vspace{.5em}
\noindent\textbf{Datasets.}
For each face-swapping method, the source and target face images are randomly selected from a pool of 30,000 identities (IDs) within the CelebA-HQ dataset~\cite{CelebAMask-HQ}. To train \name, both swapped face images and raw face images are required. As outlined in the \textbf{Note} of Section~\ref{sec:sce}, there is no overlap in identities between the training and inference phases. Specifically, IDs numbered 1 through 28,000 are allocated for the training set, while the remaining IDs are reserved for the testing set, \textcolor{black}{and the identity pool in our experiment is built up by the identity information extracted by \name from these IDs}. It is important to note that the number of images varies across different identities. 

While most contemporary face-swapping methods primarily operate at the image level as mentioned in Section~\ref{sec:fs},  we also evaluate \name's performance on swapped face video with the most commonly-used dataset FF++~\cite{ffpp} that contains videos generated from four face-swapping methods: DeepFakes~\cite{deepfakes}, FaceSwap~\cite{faceswap}, Face2Face~\cite{face2face}, and NeuralTexture~\cite{neuraltexture}. In contrast to some other video forgery datasets~\cite{dfdc,CelebDF,dfo}, FF++ labels the identity of the source and target person used in face-swapping rather than just the binary label of real or fake. 
Moreover, none of the identities in the FF++ dataset overlap with those in the CelebA-HQ dataset. Therefore, evaluation on FF++ can demonstrate transferablity of \name.

Table~\ref{tab:dataset} provides a detailed breakdown of the dataset construction for \name, showing how each face-swapping method contributes to the dataset. In Samples column, we list the number of swapped content+raw content. \textcolor{black}{As we mentioned in Section~\ref{sec:defender}, \name should extract the
correct identity information from un-swapped faces, thus we add some raw content into the test set, which are selected from those identities that produced the swapped face images in the test set.}

\begin{table}
\caption{Details of dataset construction used in \name.}
\label{tab:dataset}
\centering
\vspace{-1em}
\begin{tabular}{c|c|c|c}
\hline
\textbf{Method}              & \textbf{Format}        & \textbf{Dataset} & \textbf{\#Samples}   \\ \hline
\multirow{2}{*}{HiRes~\cite{HiRes}}       & \multirow{2}{*}{Image} & Train            & 113284+27816 \\ \cline{3-4} 
                             &                        & Test             & 2350+100       \\ \hline
\multirow{2}{*}{FaceShifter~\cite{faceshifter}} & \multirow{2}{*}{Image} & Train            & 184985+27816 \\ \cline{3-4} 
                             &                        & Test             & 2675+100       \\ \hline
\multirow{2}{*}{SimSwap~\cite{simswap}}     & \multirow{2}{*}{Image} & Train            & 273875+27816 \\ \cline{3-4} 
                             &                        & Test             & 3163+100       \\ \hline
\multirow{2}{*}{InfoSwap~\cite{infoswap}}    & \multirow{2}{*}{Image} & Train            & 271115+27816 \\ \cline{3-4} 
                             &                        & Test             & 3246+100       \\ \hline
MegaFS~\cite{megafs}                       & Image                  & Test             & 2294+100       \\ \hline
DiffSwap~\cite{diffswap}                     & Image                  & Test             & 2493+100       \\ \hline
FF++~\cite{ffpp}                         & Video                  & Test             & 2000+1000      \\ \hline
\end{tabular}
\end{table}

\vspace{.5em}
\noindent\textbf{The Baseline and Evaluation Metrics.}
We utilized the most commonly used identity information extractor, ArcFace\cite{arcface}, with a backbone of ResNet18 trained on MS1M~\cite{ms1m} dataset as our baseline \textbf{B}. Since \name could be treated as a face identification system, we use the \textbf{Top-\textit{k} Accuracy (Top-\textit{k} ACC)} as the evaluation metric. This metric denotes the rate at which the correct label is among the top \textit{k} labels predicted (ranked by similarity scores) by the face identification model. A higher Top-\textit{k} ACC means that the extracted identity information can more effectively trace the source person.

\vspace{.5em}
\noindent\textbf{Implementation Details.}
We have implemented \name on Pytorch platform and trained the model with a single NVIDIA A6000 GPU. Each input images and reference images are cropped and aligned to the size of $112\times 112$ with the pre-trained MTCNN~\cite{mtcnn} model in both training and inference phases. The settings of the identity information extraction model and the loss function were listed in  Section~\ref{sec:IIE} and Section~\ref{sec:IID}. During the training phase, we use the Ranger optimizer~\cite{Ranger} with the parameters $(0.95,0.999)$ and the Cosine learning rate scheduler to update the parameters of the model for 20 epochs, with an initialized learning rate of 1e-4, a weight decay of 2e-5, and a batch size of 64. The total training process of \name takes around 24 to 28 hours, depending on the magnitude of the data in the training set.

\subsection{Overall Evaluation}

\begin{table}[t]
\caption{Performance comparison between the baseline and \name under different scenarios.}
\label{tab:main}
\centering
\begin{tabular}{cccccc}
\hline
Method & Metric(\%, $\uparrow$) & B & S1 & S2 & S3 \\
\hline 
\multirow{3}*{HiRes~\cite{HiRes}} & Top-1 ACC & 0.04 & 98.80 & 99.27 & 93.34 \\
~ & Top-5 ACC & 0.21 & 99.27 & 99.61 & 97.31 \\
~ & Top-10 ACC & 0.46 & 99.31 & 99.61 & 97.73 \\
\hline
\multirow{3}*{FaceShifter~\cite{faceshifter}} & Top-1 ACC & 0.03 & 99.40 & 99.40 & 99.40 \\
~ & Top-5 ACC & 0.22 & 99.55 & 99.55 & 99.55 \\
~ & Top-10 ACC & 0.41 & 99.55 & 99.55 & 99.55 \\
\hline
\multirow{3}*{SimSwap~\cite{simswap}} & Top-1 ACC & 0.03 & 99.13 & 99.19 & 98.73 \\
~ & Top-5 ACC & 0.40 & 99.41 & 99.47 & 99.16 \\
~ & Top-10 ACC & 0.70 & 99.41 & 99.50 & 99.25 \\
\hline
\multirow{3}*{InfoSwap~\cite{infoswap}} & Top-1 ACC & 0.09 & 97.97 & 98.19 & 91.83 \\
~ & Top-5 ACC & 0.31 & 98.70 & 98.79 & 96.36 \\
~ & Top-10 ACC & 0.63 & 98.79 & 98.89 & 97.34 \\
\hline
\end{tabular}
\end{table}

\begin{figure}
\centering
\includegraphics[width=9.5cm]{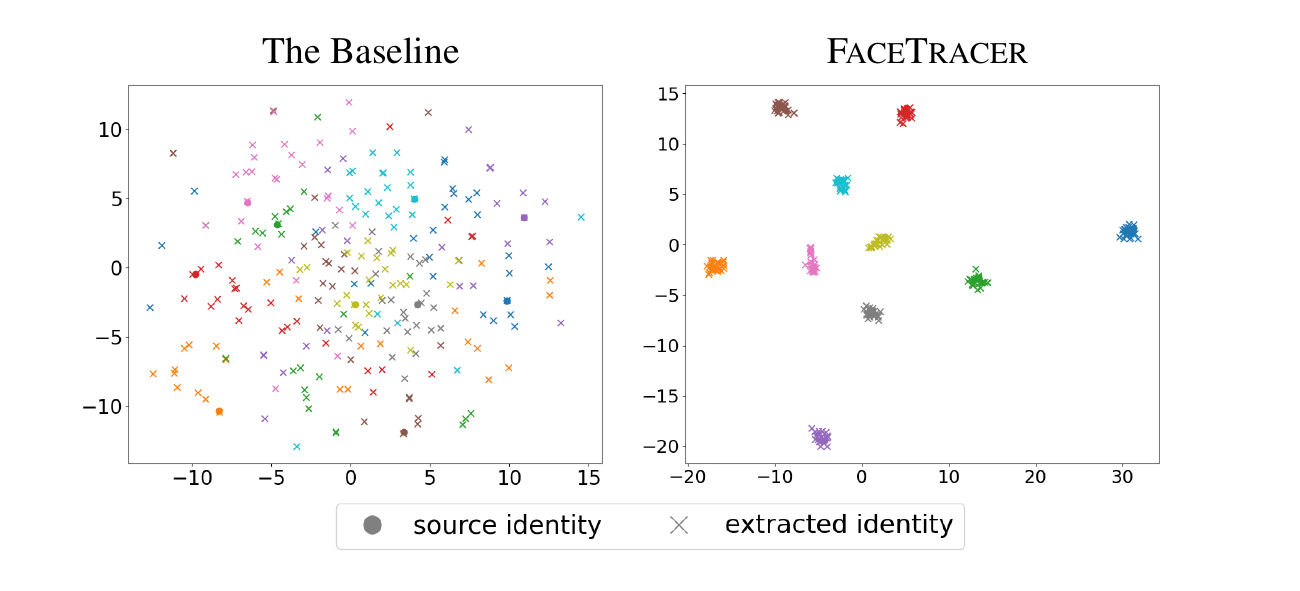}
\vspace{-2em}
\caption{T-SNE visualization of the ground-truth identity information of the source person and the identity information extracted by the baseline (left) and \name (right).}
\label{fig:tsne-compare}
\vspace{-0.5em}
\end{figure}

\begin{figure}
\centering
\includegraphics[width=9cm]{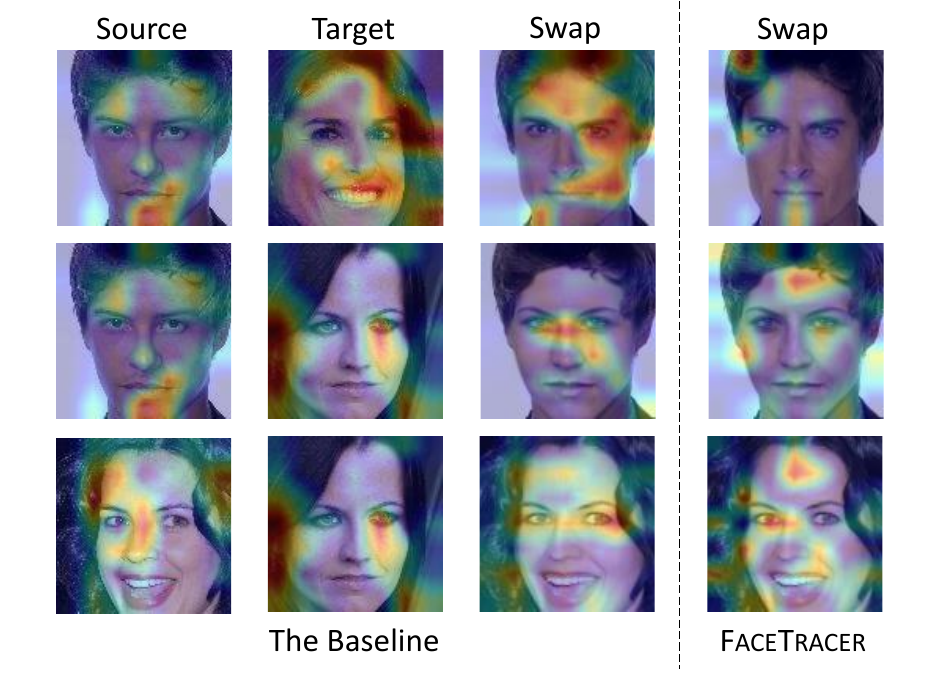}
\vspace{-1.5em}
\caption{Saliency map visualization of the images during the face-swapping process from the Baseline (left) and \name (right). \name focuses on regions such as hair, face shape, cheeks, bridge and forehead to extract the identity information of the source person.}
\label{fig:heatmap}
\vspace{-0.5em}
\end{figure}

\begin{figure*}
\centering
\includegraphics[width=17cm]{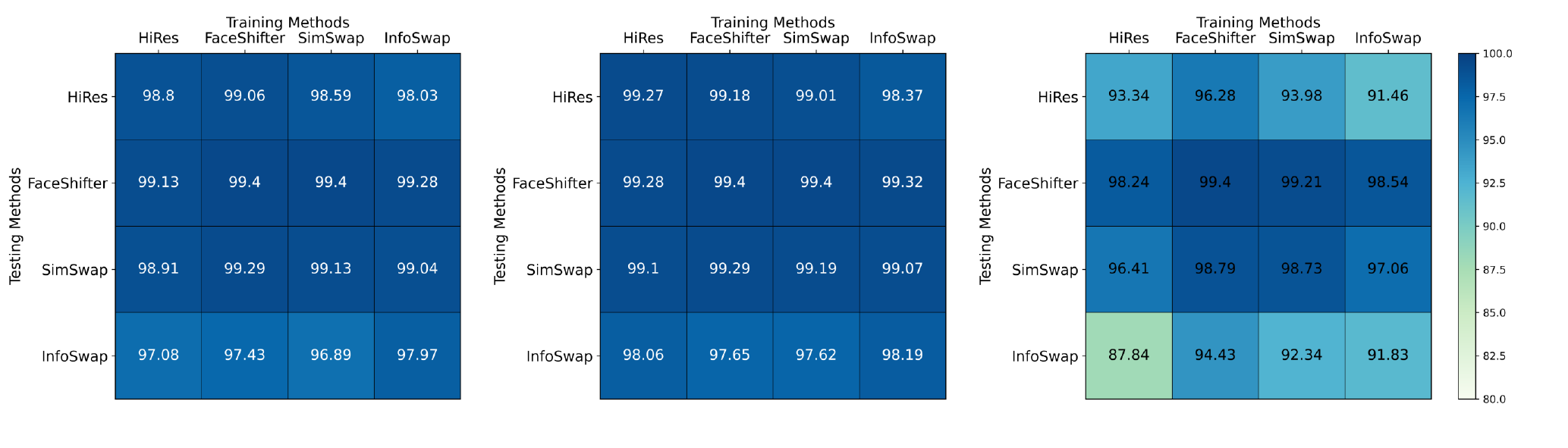}
\vspace{-0.5em}
\caption{Transferability of \name to different face-swapping methods that \textbf{explicitly} disentangle identity and attribute information, different figures represent different scenarios. Left: full-reference scenario (S1); Middle: half-reference scenario (S2); Right: none-reference scenario (S3). We adopt Top-1 ACC ($\uparrow$) as the metrics.  \name exhibits strong transferability across these methods. }
\label{fig:generalization-1-top1}
\end{figure*}

\begin{figure*}
\centering
\includegraphics[width=17cm]{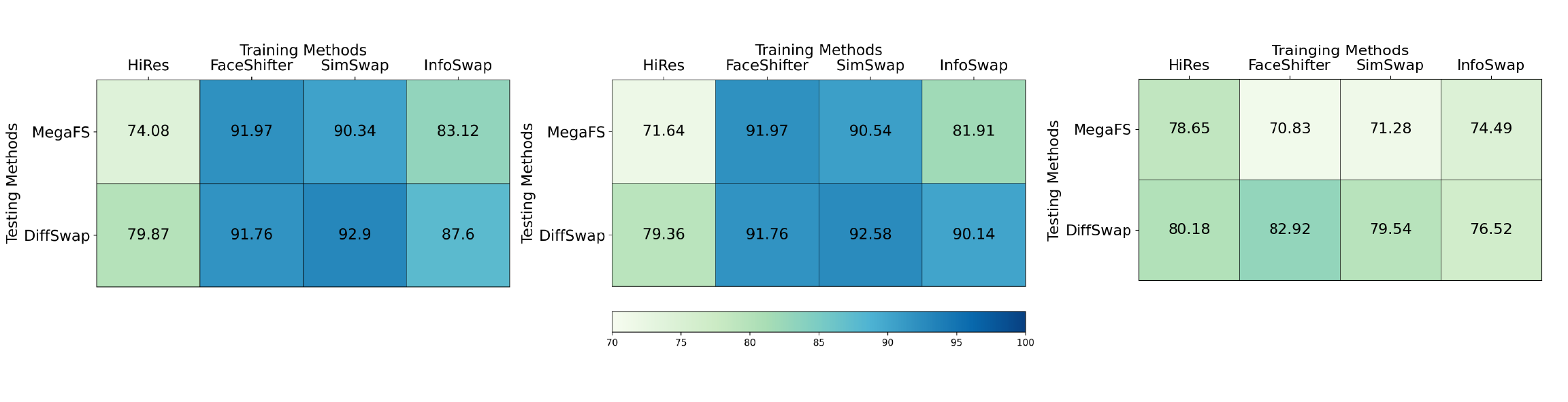}
\vspace{-0.5em}
\caption{Transferability of \name to face-swapping methods that \textbf{implicitly} disentangle identity and attribute information, different figures represent different scenarios. Left: Full-reference scenario (S1); Middle: Half-reference scenario (S2); Right: None-reference scenario (S3). We adopt Top-1 ACC ($\uparrow$) as the metrics. \name still exhibits transferability across these methods.}
\label{fig:generalization-2-top1}
\end{figure*}

In this part, we evaluated the effectiveness of \name in extracting identity information of the source person in swapped face images. As mentioned in \Sref{sec:sce}, we set up three different scenarios, namely, the full-reference scenario (S1), the half-reference scenario (S2), and the none-reference scenario (S3). Here, the training set and inference set are built on the same face swapping methods.

As shown in Table~\ref{tab:main}, \name achieves an high top-1 accuracy under different scenarios, while the top-10 accuracy of the baseline is no more than 1\%. This demonstrates that \name can \textbf{effectively} trace the source person in suspicious images with a high accuracy, regardless of whether the suspicious image has undergone face-swapping, while the baseline method fails in all cases. We explain that the baseline tends to extract the identity information relevant to the target person in the image rather than the source person. In contrast, \name mitigates the influence of the target person's identity information, enabling the accurate extraction of the identity of the source person. Noticeably, there is negligible difference between S1 and S2, but both outperform S3, indicating that using reference images during the training phase can enhance \name's performance.

To better depict the effectiveness of our framework, we presented the t-SNE analysis results of the baseline and \name in Figure~\ref{fig:tsne-compare}. We demonstrate this result using \name trained on the SimSwap method with the S1 setting as an example. Initially, we randomly selected 10 facial images of different people as the source faces and applied the SimSwap method to these images with different target faces which are also randomly selected. Subsequently, we extracted the identity information from these swapped face images with both the baseline method and \name. The results indicate that \name can accurately extract the identity information of the source individual from the input images, whereas traditional identity information extraction networks fail. 

To better understand \name's ability to trace source identities, we use saliency maps to visualize what region the model focuses on. As illustrated in Figure~\ref{fig:heatmap}, when processing swapped face images, the baseline model tends to focus on regions that are similar to those of the target image. This bias leads the baseline to identify these images as the target person, which is the training objective of the face swapping method. In contrast, \name effectively detects regions that are more relevant to recognizing the source identity, thereby enabling the extraction of source-related identity information for accurate tracing. Please note that \name places greater emphasis on regions such as hair, face shape, cheeks, bridge and forehead, these regions contain identity information of the source person even after face swapping algorithms are applied, as these regions are slightly changed after the face swapping process.

Additionally, we shall point out that it takes less than 0.05 seconds for \name to extract identity information of the source person from the input image, which shows the efficiency of \name.

\subsection{Transferability}
\label{sec:gen}

In this part, we first evaluated the transferability of \name across four face-swapping methods that explicitly disentangle identity and attribute information, where each model was trained on one face-swapping method and tested on the other three face-swapping methods. We conducted evaluations under S1, S2, and S3 scenarios. 
As shown in Figure~\ref{fig:generalization-1-top1}, \name holds strong transferability across these four face-swapping methods. 
Furthermore, the transferability evaluations conducted on \name show that using reference images during the training phase can enhance \name's transferability to some extent. For instance, when tested on swapped face images generated by InfoSwap~\cite{infoswap}, \name trained on HiRes under S1 and S2 scenarios 
achieve 97.08\% and 98.06\% Top-1 ACC, while the \name trained under S3 scenario only obtains 87.84\% Top-1 ACC.

Moreover, we also conducted experiments on the transferability of \name in handling swapped face images generated by methods that do not explicitly disentangle identity and attribute information. Figure~\ref{fig:generalization-2-top1} shows that the performance of \name decreases on the swapped face images generated by these methods compared with the previous four face-swapping methods, but still maintains an acceptable level. Although these two methods do not explicitly disentangle identity and attribute information, the identity information of the source person still partially remains in the swapped face images. For instance, MegaFS~\cite{megafs} uses the deep StyleGAN latent codes of the source image as the latent codes for generating swapped face images, and DiffSwap~\cite{diffswap} uses the noise image obtained by adding noise to the source image as the starting point for generating swapped face images. Therefore, \name still exhibits transferability to these methods. Additionally, we also observed an interesting point: when testing the transferability performance on MegaFS, there was a greater decrease in performance compared to other methods. This may be because MegaFS directly manipulates latent codes without extracting identity or attribute information. Figure~\ref{fig:generalization-1-top1} and Figure~\ref{fig:generalization-2-top1} only provide the Top-1 ACC results, and more results (Top-5$\&$Top-10) can be seen in the supplementary material, which share a consistent conclusion.

We conducted experiment to simulate the situation of using \name without knowledge of the face swapping method, we randomly selected 1,000 images generated by the aforementioned six face swapping methods and tested the performance of \name trained on HiRes data, which achieved 92.78\% Top-1 accuracy under S1 scenario.

\noindent\textbf{Commercial Apps.} For better demonstration of the performance of \name, we generate 50 swapped-face images with two commercial face swapping apps, Faceover~\cite{faceover} and DeepFaker~\cite{deepfaker}, and \name demonstrates 100\% and 96\% tracing accuracy in average, respectively.

\begin{table*}[t]
\caption{Performance comparison of \name under Intra- \& Inter-gender face swapping (source$\leftarrow$target)}
\label{tab:gender}
\vspace{-1em}
\centering
\begin{tabular}{ccccccccc}
\hline
\multirow{2}*{Train} & \multirow{2}*{Metric(\%, $\uparrow$)} & \multicolumn{5}{c}{Test(S1/S2/S3)} \\
~ & ~ & All & M$\leftarrow$M & F$\leftarrow$F & M$\leftarrow$F & F$\leftarrow$M \\
\hline
\multirow{3}*{HiRes~\cite{HiRes}} & Top-1 ACC & 98.80/99.27/93.34 & 98.57/99.40/92.26 & 99.09/99.24/95.22 & 98.75/98.75/86.25 & 97.11/99.03/83.65 \\
~ & Top-5 ACC & 99.27/99.61/97.31 & 98.92/99.64/97.38 & 99.62/99.69/98.03 & 98.75/98.75/95.00 & 98.07/99.03/89.42 \\
~ & Top-10 ACC & 99.31/99.61/97.73 & 99.04/99.64/97.61 & 99.62/99.69/98.40 & 98.75/98.75/96.25 & 98.07/99.03/91.34 \\
\hline
\multirow{3}*{FaceShifter~\cite{faceshifter}} & Top-1 ACC & 99.40/99.40/99.40 & 99.65/99.65/99.65 & 99.40/99.40/99.40 & 98.95/98.95/98.95 & 98.07/98.07/98.07 \\
~ & Top-5 ACC & 99.55/99.55/99.55 & 99.65/99.65/99.65 & 99.66/99.66/99.66 & 98.95/98.95/98.95 & 98.07/98.07/98.07 \\
~ & Top-10 ACC & 99.55/99.55/99.55 & 99.65/99.65/99.65 & 99.66/99.66/99.66 & 98.95/98.95/98.95 & 98.07/98.07/98.07 \\
\hline
\multirow{3}*{SimSwap~\cite{simswap}} & Top-1 ACC & 99.13/99.19/98.73 & 99.79/99.79/99.42 & 99.03/99.09/98.79 & 98.71/99.03/96.62 & 97.97/97.97/98.07 \\
~ & Top-5 ACC & 99.41/99.47/99.16 & 99.79/99.79/99.58 & 99.45/99.57/99.33 & 99.35/99.35/97.29 & 97.97/97.97/98.71 \\
~ & Top-10 ACC & 99.41/99.50/99.25 & 99.79/99.79/99.58 & 99.45/99.63/99.45 & 99.35/99.35/97.63 & 97.97/97.97/98.71 \\
\hline
\multirow{3}*{InfoSwap~\cite{infoswap}} & Top-1 ACC & 97.97/98.19/91.83 & 97.83/98.14/92.25 & 98.12/98.37/92.18 & 97.97/97.97/88.21 & 97.50/97.50/91.55 \\
~ & Top-5 ACC & 98.70/98.79/96.36 & 98.34/98.45/95.76 & 99.00/99.06/96.62 & 98.64/98.98/95.71 & 98.21/98.21/97.29 \\
~ & Top-10 ACC & 98.79/98.89/97.34 & 98.45/98.55/96.79 & 99.06/99.18/97.56 & 98.98/98.98/97.14 & 98.21/98.21/97.97 \\
\hline
\end{tabular}
\vspace{-1em}
\end{table*}

\begin{figure*}[t]
\centering
\includegraphics[width=17cm]{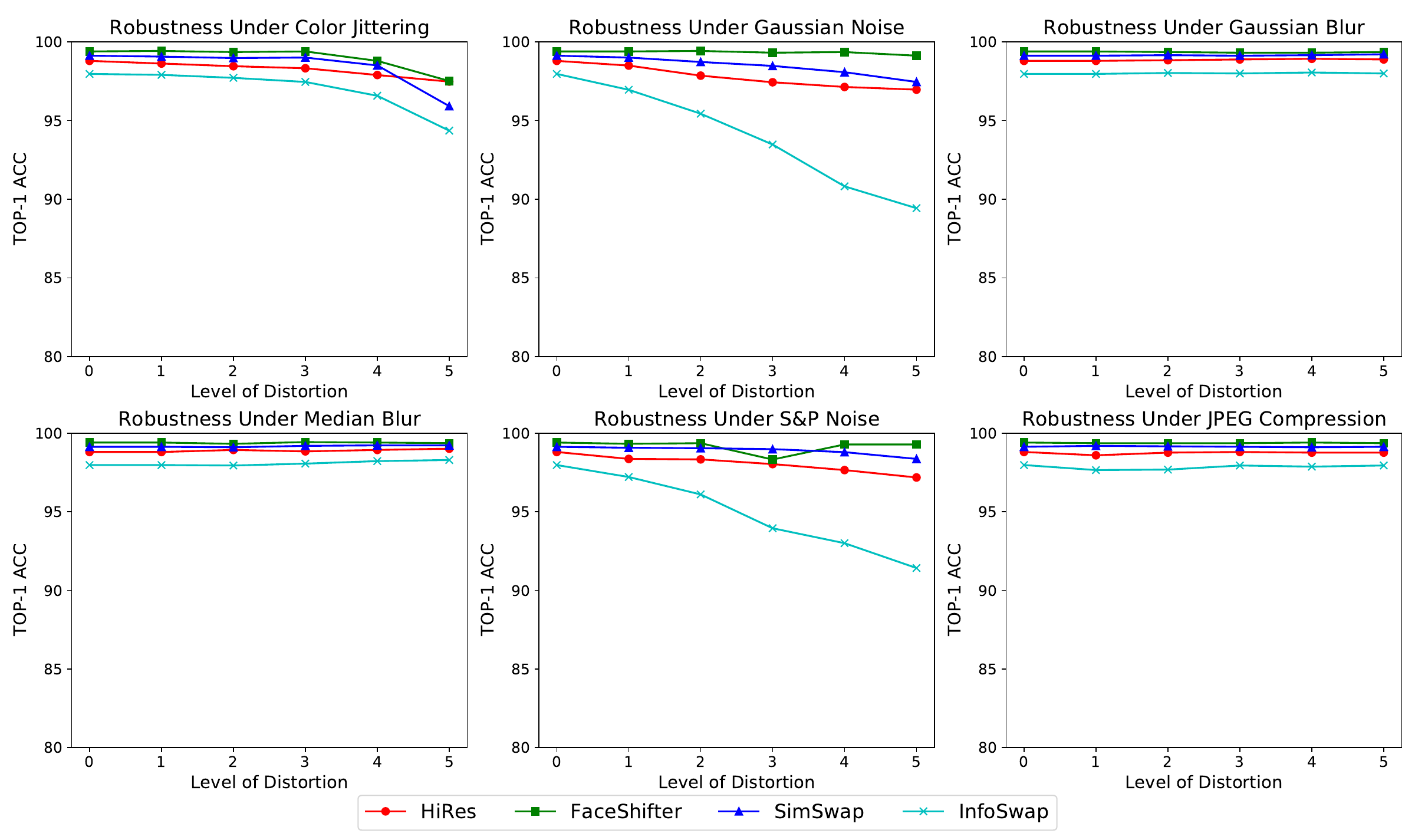}
\vspace{-1em}
\caption{
\name holds excellent robustness under various distortions that may occur during the network transmission. }
\label{fig:robustness}
\vspace{-0.8em}
\end{figure*}

\subsection{Intra- \& Inter-gender Performance}

In this part, we measured the performance of \name when dealing with different source and target genders. According to our experience, the selection of source and target people can affect the quality of swapped face images. Therefore, we divided the testing sets of each face-swapping method into four parts and tested the face identification performance of \name under different scenarios. Here, M$\leftarrow$F indicates that the source person is male and the target person is female, the other notations follow the same logic. The results are listed in Table~\ref{tab:gender}. It can be observed that \name does not lose face identification performance in almost every subsets under different scenarios, although some visually unfavorable swapped face images may be contained in the subsets.

\subsection{Robustness} \label{sec:robustness}

In this part, we investigated the robustness of \name, \ie, its ability to handle various distortions that may occur during network transmission. We applied some common distortions to the testing images, including Gaussian noise, salt-and-pepper noise, Gaussian blur, median blur, JPEG compression, and color jittering. For each type of distortion, we set five levels to study the impact of increasing distortion on the performance of \name. We listed the parameters of different levels of distortions in the supplemetary material. Taking S1 scenario as an example (see Figure~\ref{fig:robustness}), \name exhibits outstanding robustness when dealing with Gaussian blur, median blur, and JPEG compression. Additionally, although the performance of \name decreases when facing the distortions like Gaussian noise, salt-and-pepper noise, and color jittering, it is still acceptable even under extreme distortions.

\subsection{Extension to Swapped Face Videos}
\label{sec:video}

\begin{figure}
\centering
\includegraphics[width=7cm]{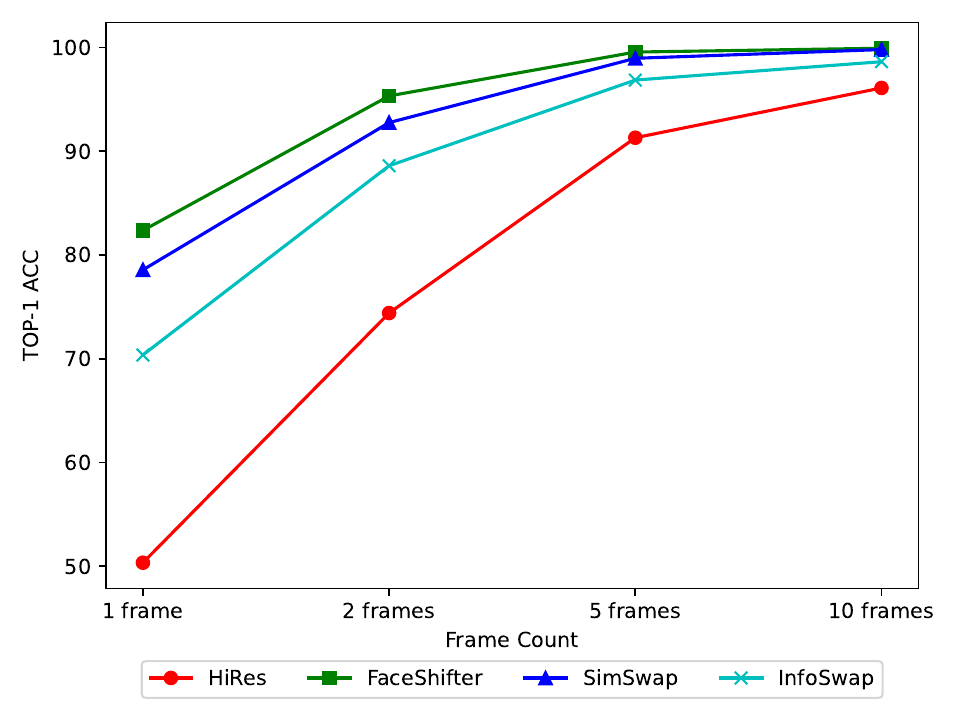}
\vspace{-1em}
\caption{Performance of \name when extents to swapped face videos under S1 scenario. Although \name performs not very effective with 1 frame input due to the video quality of FF++, its performance improves significantly as the number of input frames increases. We adopt Top-1 ACC ($\uparrow$) as the metrics.}
\label{fig:ffpp-top1}
\vspace{-0.5em}
\end{figure}

Here, we conducted experiments on the FF++ dataset~\cite{ffpp} under the S1 scenario, and the results are demonstrated in Figure~\ref{fig:ffpp-top1}, where the frame count refers to the number of frames selected from the suspected video input $V$. Specifically, we randomly selected several frames from the video, and then fed the aligned images paired with the reference image $\{I_{tar}^r\}$ into \name, where $\{I_{tar}^r\}$ will be copied several times and thus paired with the selected video frames. The final identity information could be simply computed as the average of the several outputs.
 
It can be observed that increasing the number of input video frames effectively improves the performance of \name. More results could be seen in the supplementary material. This will greatly facilitates the real-world scenario, where evidence for fraud and information about the target person is more likely to appear in video format, and the defender will be able to extract more accurate identity information about the attacker. 
Additionally, it is noted that \name exhibits not-so-good transferability performance when testing on the FF++ dataset. This could be due to the low resolution of FF++ itself, and the fact that the facial region occupies only a portion of the entire video. As a result, additional interpolation operations are required for face alignment to achieve a resolution of $112\times112$, leading to its not-so-good performance. 

\begin{table}
\caption{Performance comparison of \name with different structures.}
\label{tab:ablation}
\vspace{-1em}
\centering
\begin{tabular}{cccccc}
\hline
Method & Metric(\%, $\uparrow$) & M1 & M2 & M3 & M4\\
\hline
\multirow{3}*{HiRes~\cite{HiRes}} & Top-1 ACC & 98.80 & 78.71 & 97.56 & 89.51\\
~ & Top-5 ACC & 99.27 & 84.51 & 98.67 & 95.17\\
~ & Top-10 ACC & 99.31 & 87.24 & 98.89 & 96.37\\
\hline
\multirow{3}*{FaceShifter~\cite{faceshifter}} & Top-1 ACC & 99.40 & 99.25 & 99.28 & 98.05\\
~ & Top-5 ACC & 99.55 & 99.48 & 99.47 & 99.02\\
~ & Top-10 ACC & 99.55 & 99.48 & 99.51 & 99.21\\
\hline
\multirow{3}*{SimSwap~\cite{simswap}} & Top-1 ACC & 99.13 & 99.23 & 99.16 & 98.39\\
~ & Top-5 ACC & 99.41 & 99.44 & 99.47 & 99.22\\
~ & Top-10 ACC & 99.41 & 99.48 & 99.47 & 99.38\\
\hline
\multirow{3}*{InfoSwap~\cite{infoswap}} & Top-1 ACC & 97.97 & 97.94 & 97.91 & 96.64\\
~ & Top-5 ACC & 98.70 & 98.54 & 98.79 & 98.10\\
~ & Top-10 ACC & 98.79 & 98.64 & 98.92 & 98.35\\
\hline
\end{tabular}
\vspace{-.5em}
\end{table}

\subsection{Ablation Studies}
\label{sec:ablation}

\noindent\textbf{\textcolor{black}{Model Structure.}} In this part, we first performed ablation studies among different structures of \name:
\begin{itemize}[leftmargin=*]
    \item M1: Default setting of \name.
    \item M2: Replacing the ViT-S structure with ResNet18 structure used in the identity information extraction module and keeping the rest part of \name unchanged.
    \item M3: Replacing the disentanglement module with a cross-attention module and keeping the rest part of \name unchanged.
    \textcolor{black}{\item M4: Replacing the identity information enhancement module with a plain softmax activation as \Eref{eq:softmax} and keeping the rest part of \name unchanged.}
\end{itemize}

\begin{figure}
\centering
\includegraphics[width=9cm]{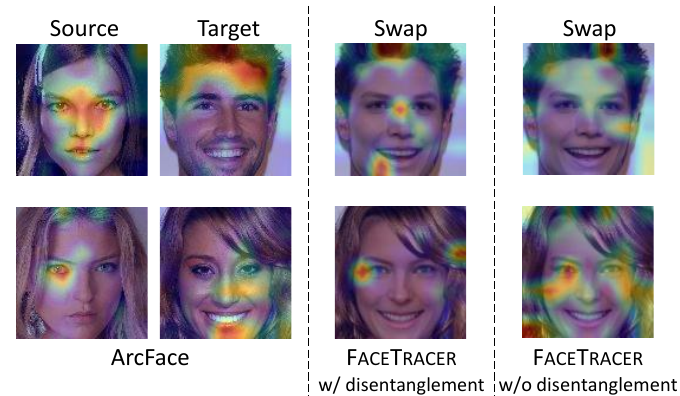}
\vspace{-1em}
\caption{Saliency map visualization of \name with and without the identity information disentanglement module. The identity information disentanglement module effectively shifts \name's focus from regions analogous to the target face to those resembling the source face.}
\label{fig:disab}
\vspace{-0.5em}
\end{figure}

\begin{figure}
\centering
\includegraphics[width=7cm]{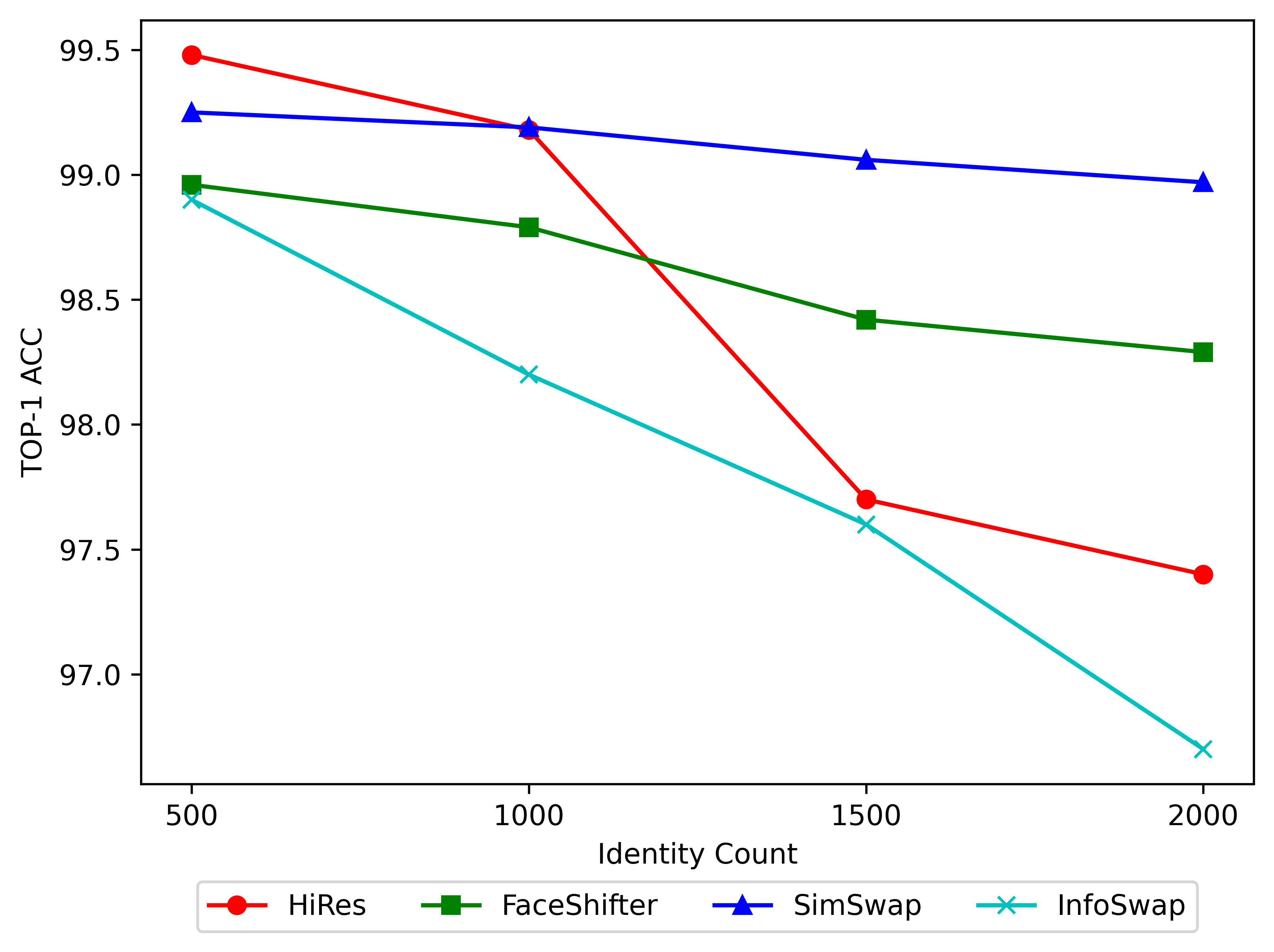}
\vspace{-1em}
\caption{Performance of \name when the size of identity pool varies. \name maintains consistent performance as the size of identity pool becomes larger. We adopt Top-1 ACC ($\uparrow$) as the metrics.}
\label{fig:pool}
\vspace{-0.5em}
\end{figure}

Under S1 scenario, we can observe from Table~\ref{tab:ablation} that different model architectures (M1, M2 and M3) do not significantly affect the performance of \name, \textcolor{black}{while changing the identity information enhancement module (M1 and M4) will affect \name's performance under each training set. This demonstrates the flexibility of \name's model structure and the effectiveness of the identity information enhancement module.}  

To demonstrate the effectiveness of the identity information disentanglement module, we used saliency maps to illustrate regions that \name focuses on with and without this module. As shown in Figure~\ref{fig:disab}, the identity information disentanglement module successfully enables the region \name focuses on switching from being similar to the target face to being similar to the source face. 

\textcolor{black}{\noindent\textbf{The Scale of the Identity Pool.} We evaluated \name's performance across varying identity pool sizes of 500, 1,000, 1,500, and 2,000 identities. For each configuration, we conducted analyses on distinct test sets, each comprising 1,000 randomly selected swapped face images, where the source identities were constrained to the respective identity pools.}

\begin{figure}
\centering
\includegraphics[width=7cm]{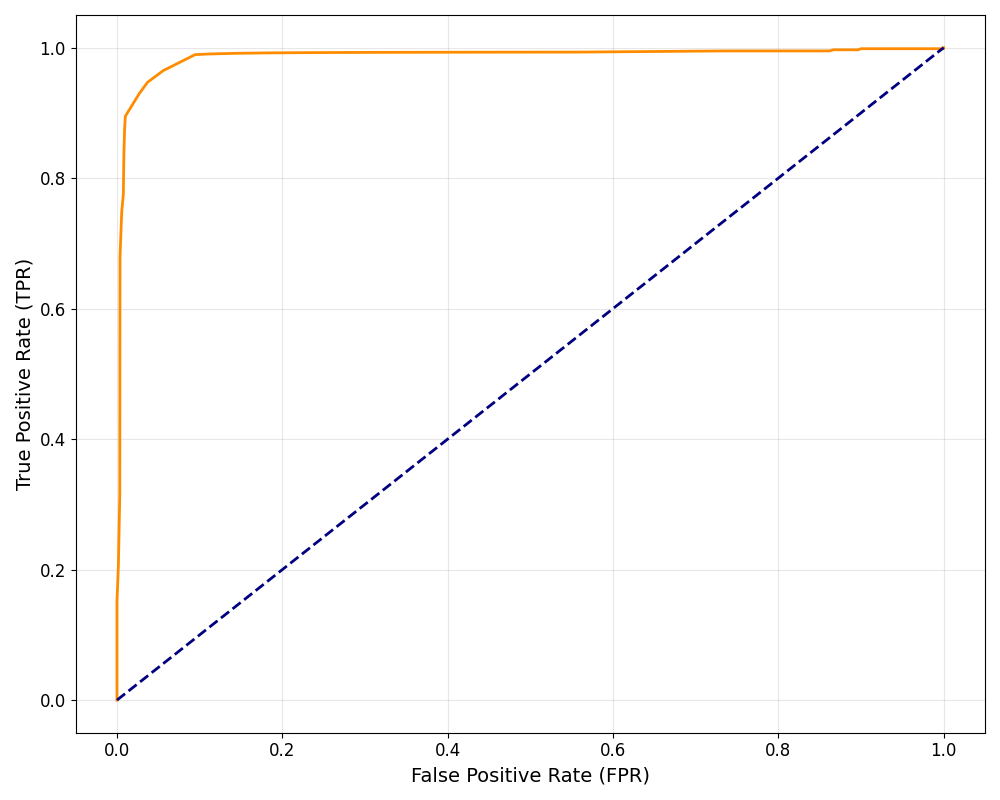}
\vspace{-1em}
\caption{Receiver Operating Characteristic (ROC) curve of the filtering strategy, this filtering strategy is simple yet effective.}
\label{fig:roc}
\vspace{-0.5em}
\end{figure}

Figure~\ref{fig:pool} illustrates that \name maintains consistent performance as the number of identities in the identity pool increases. Additionally, since \name is intended for use by trusted third parties—such as police forces and other organizations capable of maintaining large-scale identity pools—we can assume that the source identity information of input images already exists within these pools. To further optimize performance, we can initially filter out images whose source identities are not present in the identity pool by applying an identity matching threshold. We evaluated our filtering strategy through Receiver Operating Characteristic (ROC) curve analysis, presented in Figure~\ref{fig:roc}. The performance characteristics, plotted across multiple filtering threshold configurations, demonstrate the effectiveness of the strategy.

\begin{table}
\caption{Performance of \name trained on the multiple face-swapping dataset.}
\vspace{-1em}
\label{tab:ensemble}
\centering
\begin{tabular}{ccccc}
\hline
Testing Method & Metric(\%, $\uparrow$) & S1 & S2 & S3 \\
\hline
\multirow{3}*{HiRes~\cite{HiRes}} & Top-1 ACC & 99.23 & 99.36 & 93.38 \\
~ & Top-5 ACC & 99.61 & 99.65 & 97.14 \\
~ & Top-10 ACC & 99.61 & 99.65 & 97.90 \\
\hline
\multirow{3}*{FaceShifter~\cite{faceshifter}} & Top-1 ACC & 99.36 & 99.32 & 98.50 \\
~ & Top-5 ACC & 99.51 & 99.55 & 99.21 \\
~ & Top-10 ACC & 99.55 & 99.58 & 99.40 \\
\hline
\multirow{3}*{SimSwap~\cite{simswap}} & Top-1 ACC & 99.22 & 99.25 & 97.03 \\
~ & Top-5 ACC & 99.50 & 99.53 & 98.70 \\
~ & Top-10 ACC & 99.50 & 99.53 & 99.07 \\
\hline
\multirow{3}*{InfoSwap~\cite{infoswap}} & Top-1 ACC & 98.16 & 98.67 & 90.79 \\
~ & Top-5 ACC & 98.63 & 98.92 & 96.13 \\
~ & Top-10 ACC & 98.73 & 98.95 & 97.18 \\
\hline
\end{tabular}
\vspace{-1em}
\end{table}

\subsection{Ensemble Training Strategy}

We trained \name on dataset built on a single face swapping method by default. Here, we further evaluated the impact of ensemble training strategy, namely training on dataset consisting of different face-swapping methods. Specifically, we sampled one-fourth of the swapped face images from the four training sets respectively and made a training dataset with multiple face-swapping methods. Then we trained \name on this dataset. As shown in Table~\ref{tab:ensemble}, \name achieves excellent performance on different testing sets under different scenarios. Therefore, users of \name could collect training data from different face-swapping methods for training without affecting \name's performance.

\subsection{Continuous Face-swapping Attacks}
\label{sec:adaptive}

\begin{table}
\caption{Performance of \name against continuous face swapping.}
\vspace{-1em}
\label{tab:adaptive}
\centering
\begin{tabular}{cccccc}
\hline
Method & Metric(\%,$\uparrow$) & S1 & S2 & S3 \\
\hline
\multirow{3}*{HiRes~\cite{HiRes}} & Top-1 ACC & 97.85 & 98.56 & 84.73 \\
~ & Top-5 ACC & 98.75 & 99.23 & 92.41 \\
~ & Top-10 ACC & 99.09 & 99.28 & 93.79 \\
\hline
\multirow{3}*{FaceShifter~\cite{faceshifter}} & Top-1 ACC & 100.0 & 100.0 & 99.59 \\
~ & Top-5 ACC & 100.0 & 100.0 & 99.89 \\
~ & Top-10 ACC & 100.0 & 100.0 & 99.94 \\
\hline
\multirow{3}*{SimSwap~\cite{simswap}} & Top-1 ACC & 99.33 & 99.23 & 94.69 \\
~ & Top-5 ACC & 99.63 & 99.74 & 97.55 \\
~ & Top-10 ACC & 99.74 & 99.79 & 98.16 \\
\hline
\multirow{3}*{InfoSwap~\cite{infoswap}} & Top-1 ACC & 95.95 & 96.40 & 90.09 \\
~ & Top-5 ACC & 97.77 & 97.87 & 93.19 \\
~ & Top-10 ACC & 98.27 & 98.27 & 97.65 \\
\hline
\end{tabular}
\vspace{-.5em}
\end{table}

In this part, we considered adversaries who perform face-swapping on a face image with two different target faces. For example, Alice is the malicous attacker,  she uses Bob's face image as the target face image to produce swapped face image. Then, Carol's face image is further used as the second target face image to continuously produce the swapped face image, both by the same face-swapping method. In this scenario, \name needs to be able to extract Alice's identity information. In Table~\ref{tab:adaptive}, \name still achieves high accuracy on double-swapped face images, which demonstrates \name could effectively defense such adversary. In the supplementary material, we also demonstrated the ability of \name to defend against an adaptive adversary using two different face-swapping methods with two different target faces.

\begin{figure}
\centering
\includegraphics[width=8.5cm]{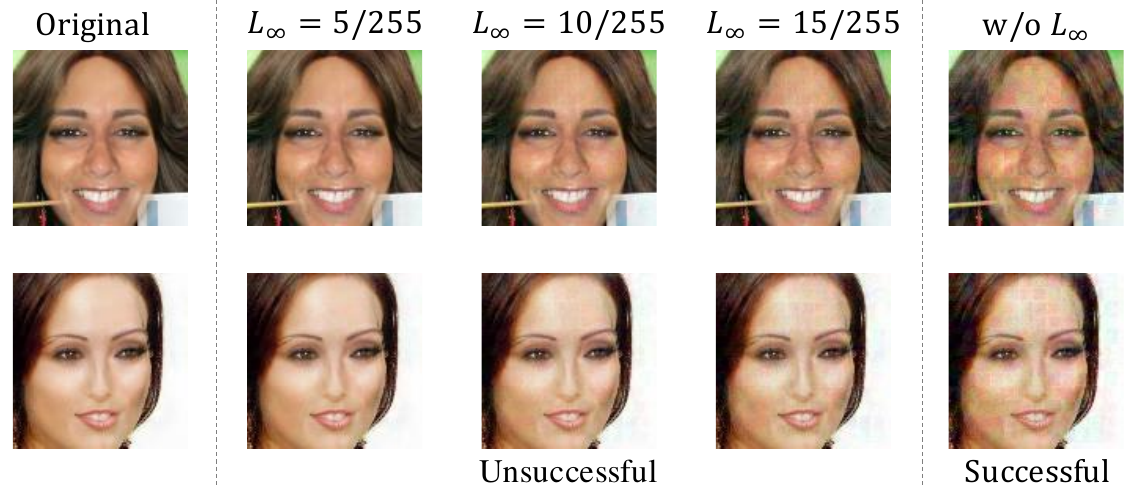}
 	\caption{Visual examples of adaptive evaluation. A successful attack, however, only occurs without $L_{\infty}$ constraints, sacrificing visual quality with clearly perceptible noise.} 
  	\label{fig:adaptive}
\end{figure}

\subsection{Adaptive Evaluation}

In this part, we considered adaptive adversaries who can access and obtain the output of \name. One intuitive adaptive attack is to circumvent \name's tracing by adding imperceptible adversarial noise to swapped face images. The goal of this attack is to alter the identity information extracted by \name so that it no longer matches the source person. To generate such adversarial noise, we utilized the widely recognized PGD~\cite{PGD} method to generate adversarial noise under different $L_{\infty}$ constraints. As shown in Figure~\ref{fig:adaptive}, the attack fails when $L_{\infty}=15/255$, where the added noise becomes clearly visible. We also present a case where the attack succeeds, but at the cost of severely compromising visual quality, making it impractical for attackers in real-world scenarios.

\section{Related Works}

\noindent\textbf{Face Swapping Techniques.}
Early face-swapping methods~\cite{realistic,dvp,neuraltexture,nirkin2018face} rely on 3D templates to disentangle identity and attribute information to process face-swapping. However, these methods lack expressiveness for some detailed features such as illumination and style.  Recently, many work introduce GANs for face swapping, such as RSGAN~\cite{rsgan}, FSNet~\cite{fsnet} and FSGAN~\cite{fsgan}. The encoder-decoder architecture is also commonly used in various face-swapping methods~\cite{faceshifter,infoswap,simswap}, which explicitly disentangles the identity of the target person and the attributes of the source person by encoding them separately with different encoders, and then uses a decoder to obtain the swapped face image. With the development of the generative models, some face-swapping methods also leverage the generative capability of the off-the-shelf models. MegaFS~\cite{megafs} and HiRes~\cite{HiRes} leverage the generative capabilities of StyleGAN~\cite{stylegan,stylegan2} to achieve face swapping, while DiffSwap~\cite{diffswap} utilizes state-of-the-art generative model, namely the diffusion models~\cite{ddpm,ddim} to accomplish face swapping.

\noindent\textbf{Face Swapping Detection.}
As the antithesis of face-swapping, face-swapping detection technology is also constantly evolving.  Early works~\cite{liy2018exposingaicreated,yang2019exposing} detect the forgery through visual biological artifacts. Some work focus on the frequency domain of the swapped face images and videos and produced methods such as~\cite{Liu_2021_CVPR,f3net} while some others focus on the temporal consistency such as~\cite{Zheng_2021_ICCV,altfreezing,tall}. Moreover, recent approach also captured precise face geometric features~\cite{facexray} or blending artifacts~\cite{sbi} to detect face-swapping images and videos. Some recent work\textcolor{black}{~\cite{dong2020identitydrivendeepfakedetection, liu2023ti2net, fan2023attacking} uses face identity information to perform face swapping detection, and}~\cite{caddm} has also found implicit identities for face-swapping results, which has somewhat inspired our approach.

\section{Discussion}

Although \name performs well as shown in the above experiments, there also exist some limitations.

\noindent\textbf{Quality of Swapped Face Images.} Since the input image size of \name is fixed, some low-resolution swapped face images may have facial parts smaller than the required input size and need scaling operations. This can lead to less accurate extraction of the source person's identity information, \ie, for low-quality swapped face data like FF++~\cite{ffpp}, \name performs unsatisfactorily with single frame input. A highly effective solution to this problem is to increase the number of input images, which can significantly enhance \name's performance (see Figure~\ref{fig:ffpp-top1}).

\noindent\textbf{Face Image Reconstruction.} Although \name extracts the identity information of the source individual, a more convenient tracing method is reconstructing the attacker's face directly from the extracted identity information using facial reconstruction techniques. This can establish an end-to-end system. We attempted to construct such a facial reconstruction network on the CelebA-HQ dataset. Unfortunately, in some attempts, this reconstruction network lacked the generalization ability to out-of-domain data. We will investigate this in future work.

\noindent\textbf{Face Swapping Methods.} In Section~\ref{sec:gen}, we found that \name exhibits a decrease in performance on swapped face images generated by MegaFS and DiffSwap. These methods, which do not explicitly disentangle identity and attribute information, retain less identity information of the source person. Nevertheless, these methods often come with drawbacks. For instance, MegaFS may easily generate meaningless images, and DiffSwap requires significant computational resources for training. In practical scenarios, traditional face-swapping methods that explicitly disentangle identity and attribute information are still predominant. Furthermore, as of now, there isn't a method that can entirely separate identity from attribute information, allowing the extraction of identity information from the source person to remain feasible.

\noindent\textbf{Privacy Enhancing Methods.} As discussed, \name could leverage privacy-preserving technologies to safeguard privacy and prevent the leakage of identity information. Fortunately, a variety of techniques are available for achieving this with large-scale facial datasets. These techniques include differential privacy~\cite{PEEP,216777}, feature subtraction~\cite{featuresub}, and adding adversarial noise~\cite{advface}. Moreover, using edge computing-based method could also be a possible option~\cite{edgecomputing}. More details can refer to the latest survey~\cite{PPFRSurvey}.

\section{Conclusion}
In this paper, we proposed the first non-intrusive tracing framework, \name, which can extract the identity information of the source person in swapped face images for effective forensics.
To achieve it, we elaborate two main modules: identity information extraction module and identity information disentanglement module.
Extensive qualitative and quantitative results demonstrate the effectiveness of \name under three practical scenarios and its
strong transferability and robustness. \name also exhibits the flexibility in model structure and can be easily extended to face-swapping videos. In the future, we consider building an end-to-end framework to reconstruct face images of the source person from swapped face images.

\bibliographystyle{IEEEtran}
\bibliography{ref}

\begin{IEEEbiography}[{\includegraphics[width=1in,height=1.25in,clip,keepaspectratio]{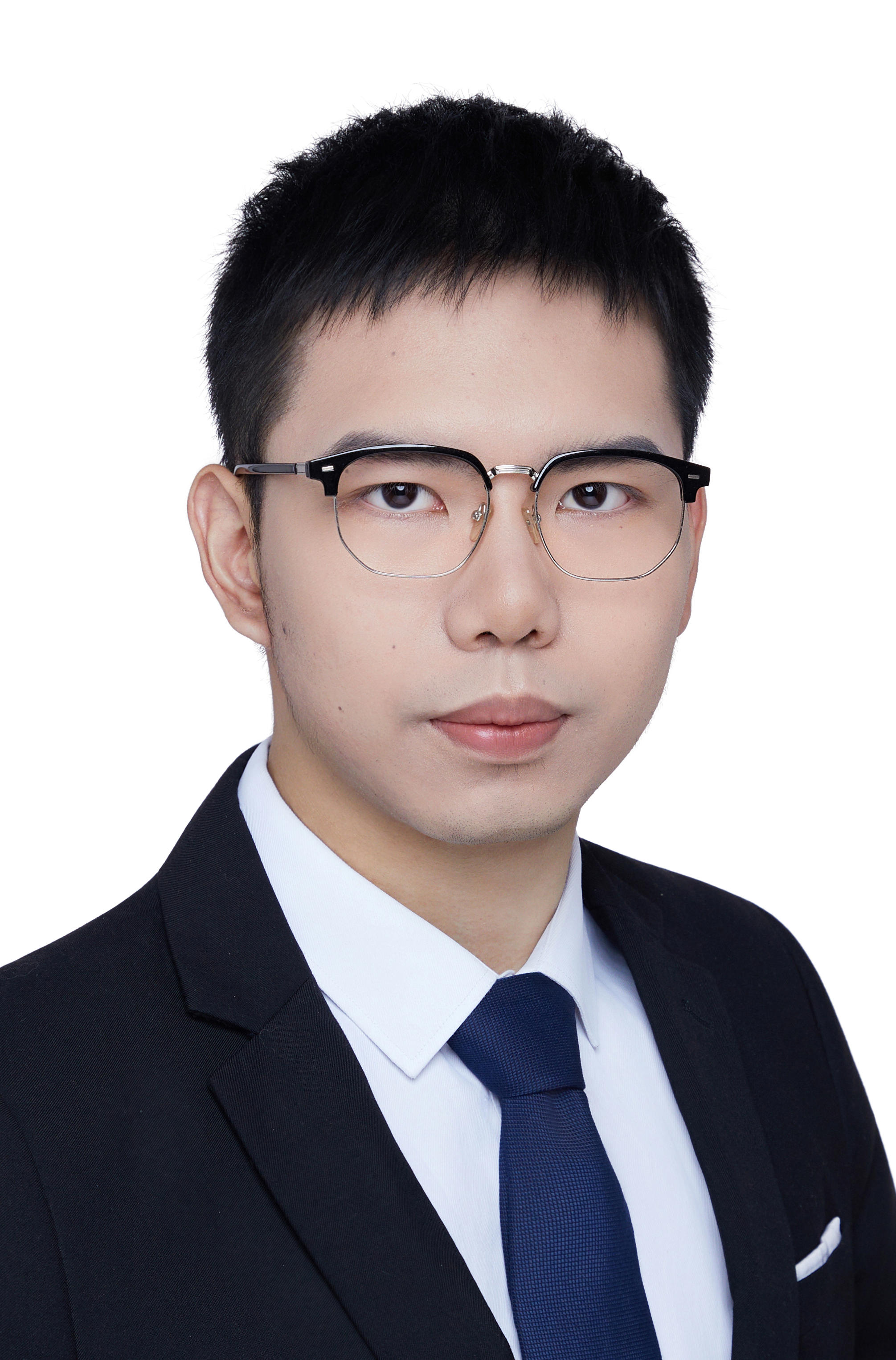}}]
{Zhongyi Zhang} is currently working toward the PhD
degree in School of Cyber Science and Technology, University of Science and Technology of China (USTC). His research interests mainly include Face Anonymization, DeepFake Detection and AIGC Generation.
\end{IEEEbiography}

\begin{IEEEbiography}[{\includegraphics[width=1in,height=1.25in,clip,keepaspectratio]{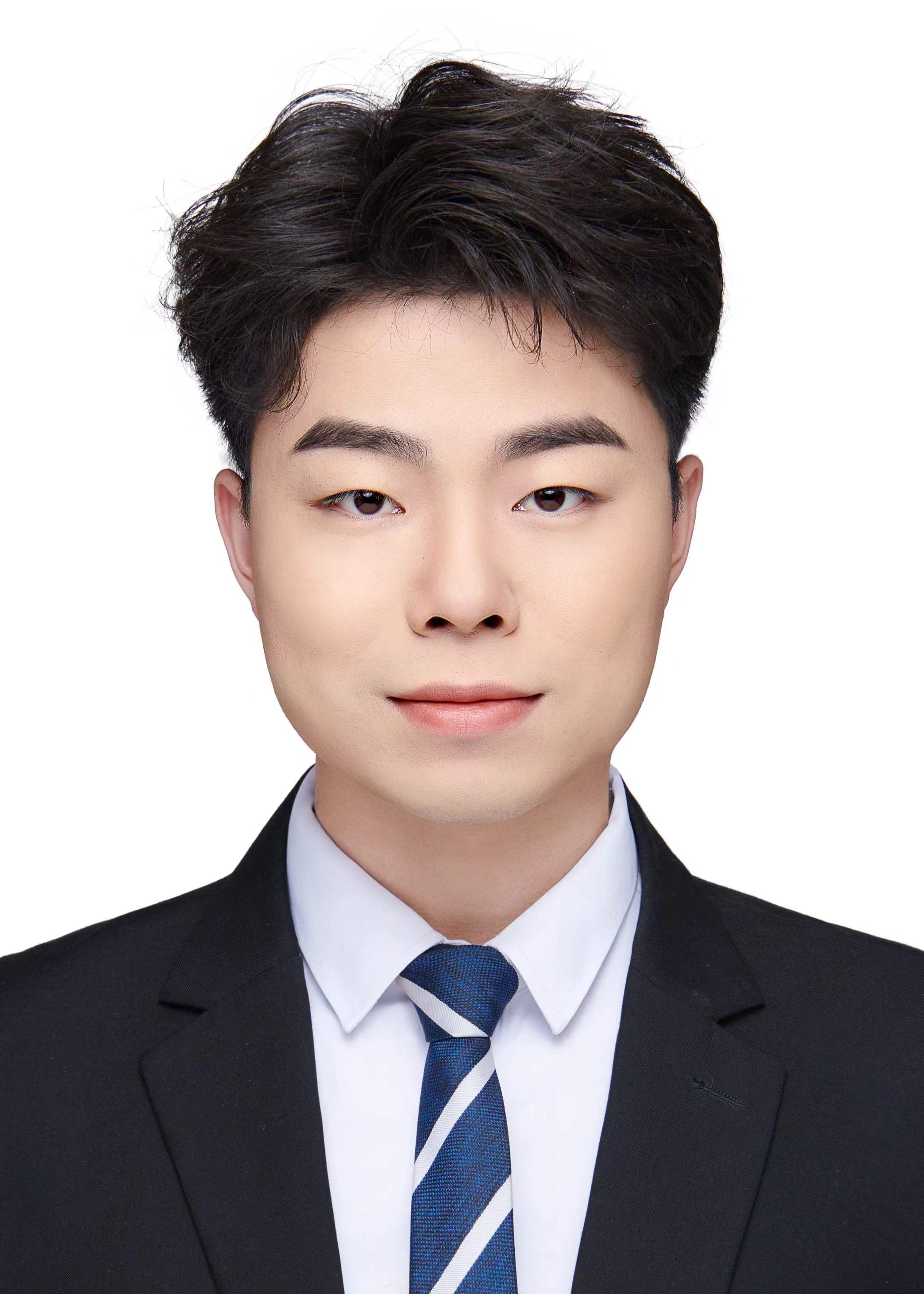}}]{Jie Zhang}
   received his B.S. degree in 2017 from
China University of Geosciences, Beijing and received his Ph.D.degree in 2022 from the University of
Science and Technology of China (USTC). Currently,
he is a Research Scientist of Center for Frontier AI Research, Agency for Science, Technology and Research (A*STAR), Singapore. His primary research interests include IP protection for AI, Trustworthy generative AI, and AI Regulation.
\end{IEEEbiography}

\begin{IEEEbiography}
[{\includegraphics[width=1in,height=1.25in,clip,keepaspectratio]{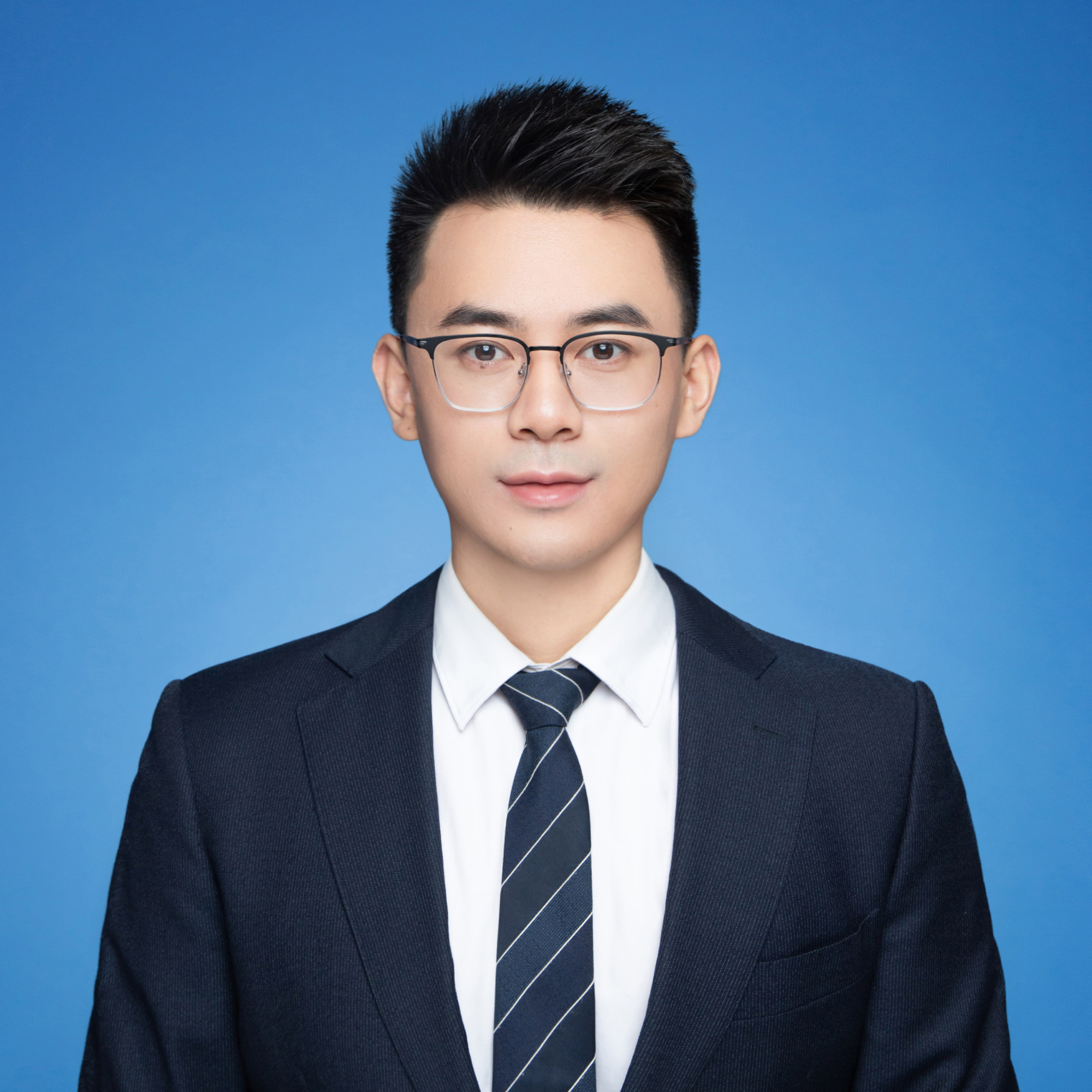}}]{Wenbo Zhou} received the BS degree from Nanjing
University of Aeronautics and Astronautics, China
in 2014, and the PhD degree from the University of Science and Technology of China in 2019, where he is currently vise professor. His research interests include information hiding and AI security.
\end{IEEEbiography}

\begin{IEEEbiography}[{\includegraphics[width=1in,height=1.25in,clip,keepaspectratio]{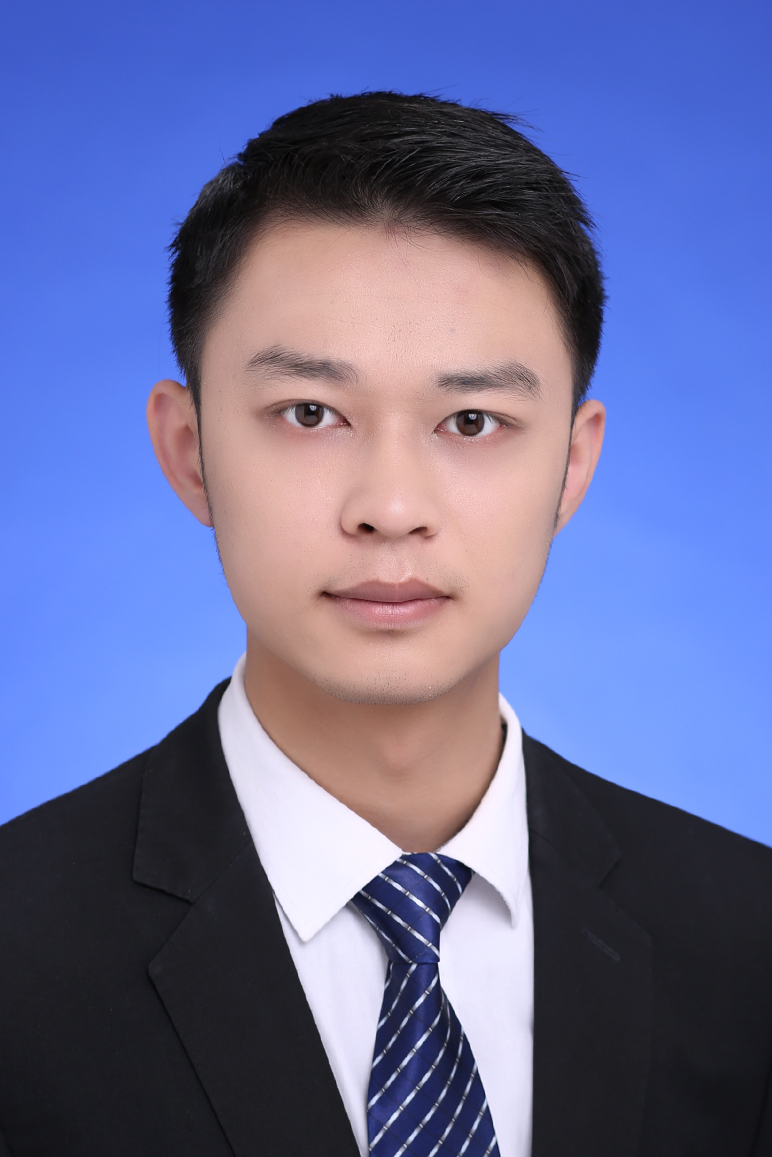}}]{Xinghui Zhou} is a PhD student at the University of Science and Technology of China in Anhui, China. He obtained his bachelor’s degree in Automation Engineering from Tianjin University and his master's degree in Computing Science from Beijing Electronic Science \& Technology Institute, where he focused on Visual Computing, Image Quality Assessment, and Artificial Intelligence.
\end{IEEEbiography}

\begin{IEEEbiography}%
[{\includegraphics[width=1in,height=1.25in,clip,keepaspectratio]{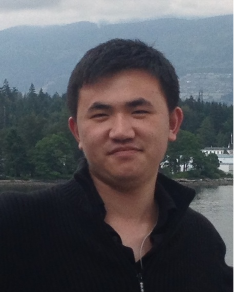}}]{Qing Guo}
received his Ph.D. degree from the School of Computer Science and Technology, Tianjin University, China. He was a research fellow and the Wallenberg-NTU Presidential Postdoctoral Fellow at the Nanyang Technological University, Singapore, from Dec. 2019 to Sep. 2022. He is currently a senior research scientist and principal investigator at the Center for Frontier AI Research (CFAR), A*STAR in Singapore. He is also an adjunct assistant professor at the National University of Singapore (NUS). He serves as the Senior PC for AAAI and Area Chair for ICLR 2025. His research mainly focuses on computer vision, AI security, adversarial attacks, and robustness. He is a member of IEEE.
\end{IEEEbiography}

\begin{IEEEbiography}%
[{\includegraphics[width=1in,height=1.25in,clip,keepaspectratio]{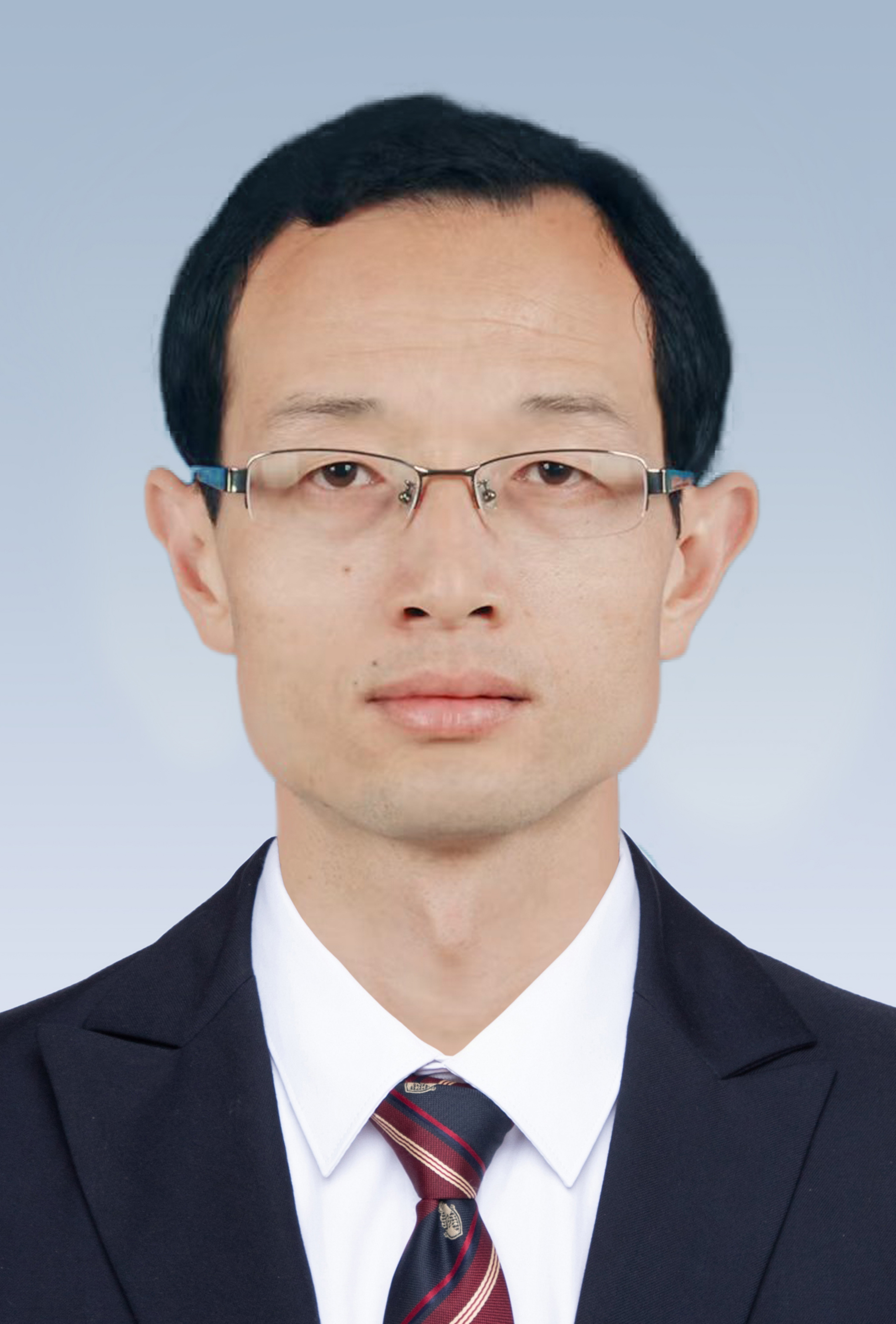}}]{Weiming Zhang} received the MS and PhD degrees
from Zhengzhou Information Science and Technology Institute, China in 2002 and 2005 respectively. Currently, he is a professor with the School of Information Science and Technology, University of Science and Technology of China. His research interests include information hiding and multimedia security.
\end{IEEEbiography}

\begin{IEEEbiography}[{\includegraphics[width=1in,height=1.25in,clip,keepaspectratio]{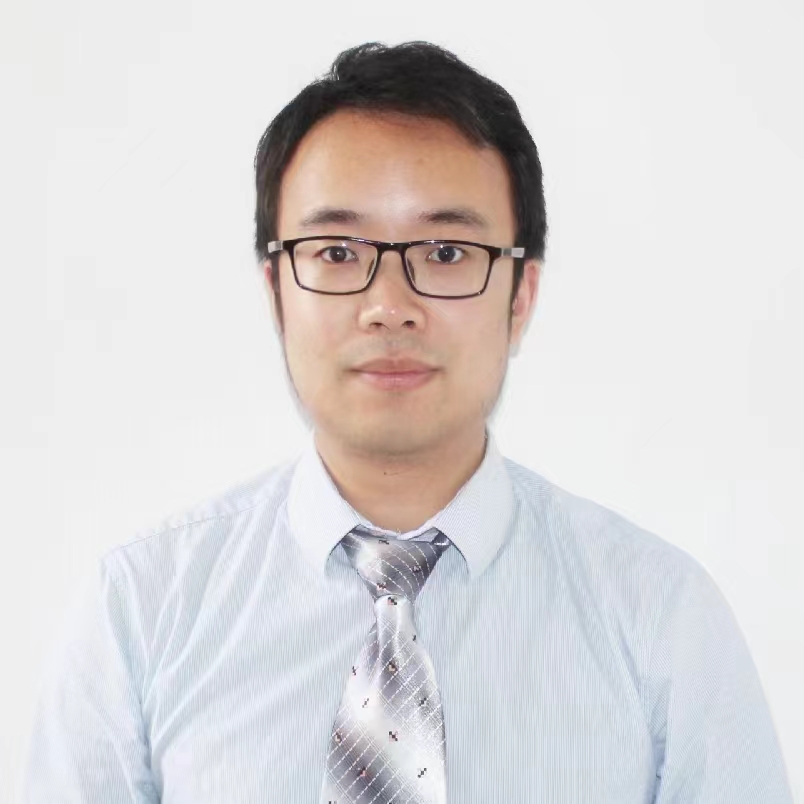}}]{Tianwei Zhang}
is an associate professor of College of Computing and Data Science at Nanyang Technological University. His research focuses on computer system security. He is particularly interested in security threats and defenses in machine learning systems, autonomous systems, computer architecture and distributed systems. He received his Bachelor's degree at Peking University in 2011, and the Ph.D. degree in at Princeton University in 2017.
\end{IEEEbiography}

\begin{IEEEbiography}
[{\includegraphics[width=1in,height=1.25in,clip,keepaspectratio]{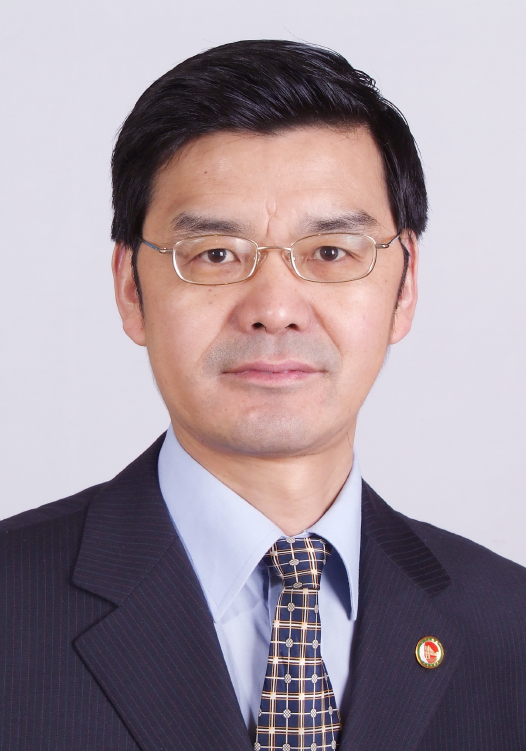}}]{Nenghai Yu} received the PhD degree from USTC in
2004. He is a full professor with the University of Science and Technology of China. He is also the director of Information Processing Center of USTC, deputy director of academic committee of School of Information Science and Technology. He was a visiting scholar in Institute of Production Technology, Faculty of Engineering, University of Tokyo, in 1999 and did
cooperative research as the senior visiting scholar in Department of Electrical Engineering, Columbia University, from Apr. to Oct. 2008. His research focuses on image processing and video analysis, multimedia communication, media content security, Internet information retrieval, data mining and content filtering, network communication and security.
\end{IEEEbiography}

\newpage

\appendices

\section{More Results}

\subsection{Feasibility of Extracting the Source Identity}
\label{app:vis}

\begin{figure}[h]
	\begin{center}
		\setlength{\tabcolsep}{0.5pt}
		\begin{tabular}{m{1.8cm}<{\centering}m{0.5cm}<{\centering}m{1.8cm}<{\centering}m{1.8cm}<{\centering}m{1.8cm}<{\centering}}						
			\includegraphics[width=1.75cm]{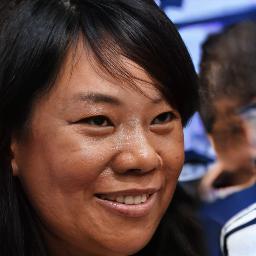}
                &
			&\includegraphics[width=1.75cm]{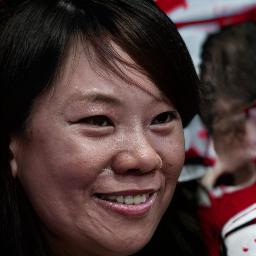}
			&\includegraphics[width=1.75cm]{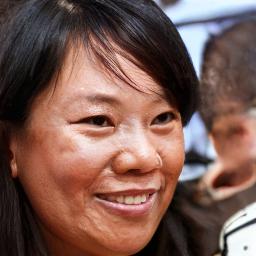}
			&\includegraphics[width=1.75cm]{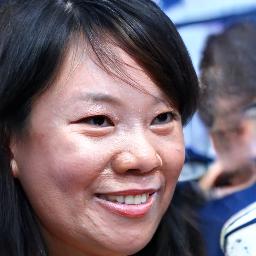}
                \\

			\includegraphics[width=1.75cm]{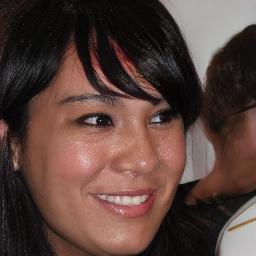}
                &
			&\includegraphics[width=1.75cm]{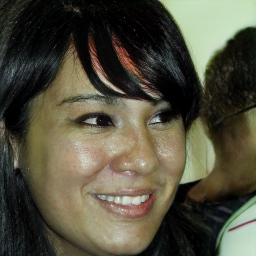}
			&\includegraphics[width=1.75cm]{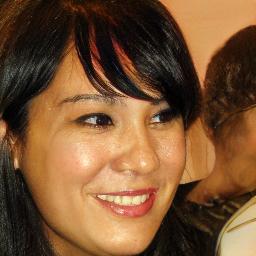}
			&\includegraphics[width=1.75cm]{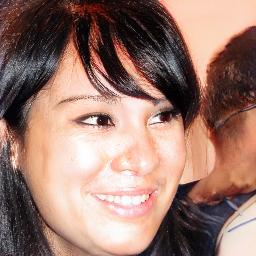}   
			\\

   			\includegraphics[width=1.75cm]{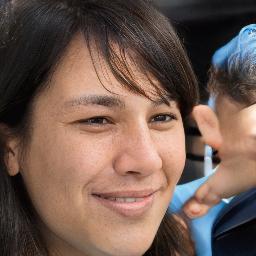}
                &
			&\includegraphics[width=1.75cm]{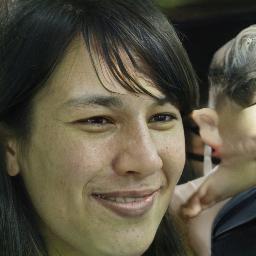}
			&\includegraphics[width=1.75cm]{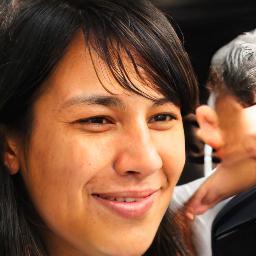}
			&\includegraphics[width=1.75cm]{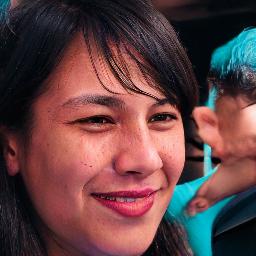}   
			\\
				
		\end{tabular}
	\end{center}
 	\caption{Visualization of the generated images. The leftmost column has the same first 12 layers of latent codes. Each row has the same latent code, which only varies in the noise inputs.} 
  	\label{fig:stylegan}
\end{figure}

In Figure~\ref{fig:stylegan}, we visualize part of the image generated in the manner mentioned in Section 2.3 of the main paper. Images generated by StyleGAN with the same first 12 layers latent codes have similar face attributes, and adjusting the noise input will generate images with almost the same attributes and identities.

\subsection{Transferability}
\label{app:gen}

\begin{figure*}[ht]
\centering
\includegraphics[width=17cm]{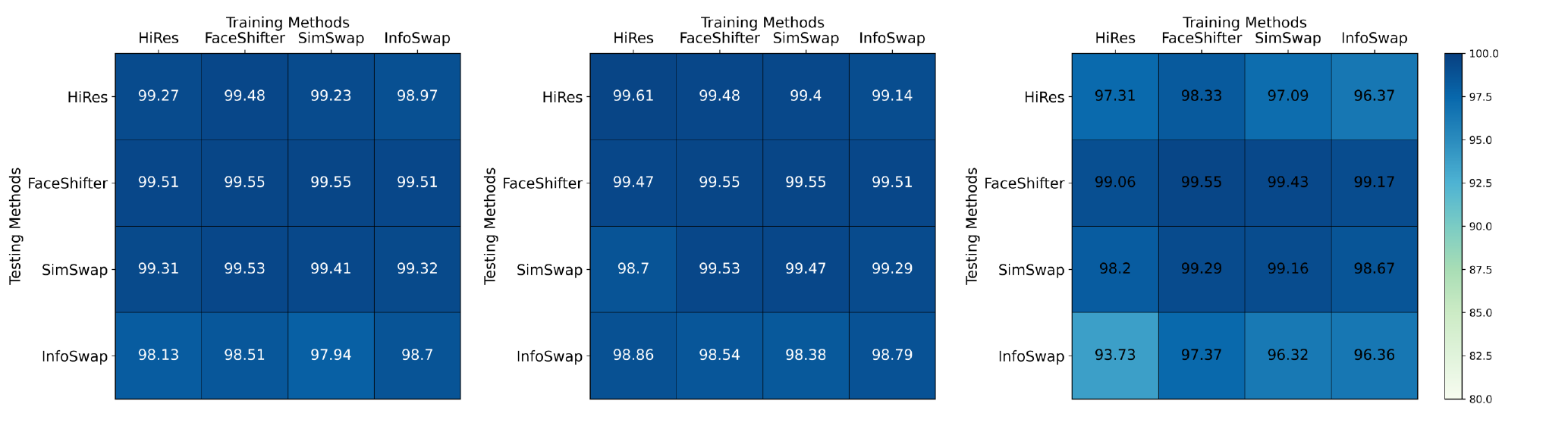}
\vspace{-1.7em}
\caption{Transferability to face-swapping methods that \textbf{explicitly} disentangle identity and attribute information, different figures represent different scenarios. Left: Full-reference scenario (S1); Middle: Half-reference scenario (S2); Right: None-reference scenario (S3). We adopt Top-5 ACC ($\uparrow$) as the metrics. \name still exhibits transferability ability on these methods.}
\label{fig:generalization-1-top5}
\end{figure*}

\begin{figure*}[ht]
\centering
\includegraphics[width=17cm]{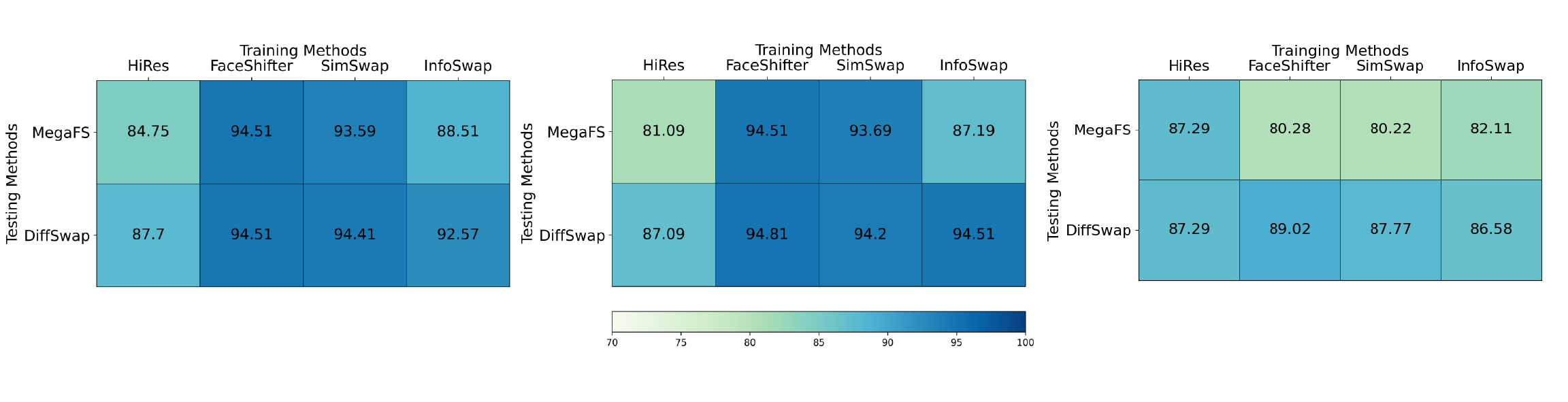}
\vspace{-2em}
\caption{Transferability to face-swapping methods that \textbf{implicitly} disentangle identity and attribute information, different figures represent different scenarios. Left: Full-reference scenario (S1); Middle: Half-reference scenario (S2); Right: None-reference scenario (S3). We adopt Top-5 ACC ($\uparrow$) as the metrics. \name still exhibits transferability on these methods.}
\label{fig:generalization-2-top5}
\end{figure*}

\begin{figure*}[ht]
\centering
\includegraphics[width=17cm]{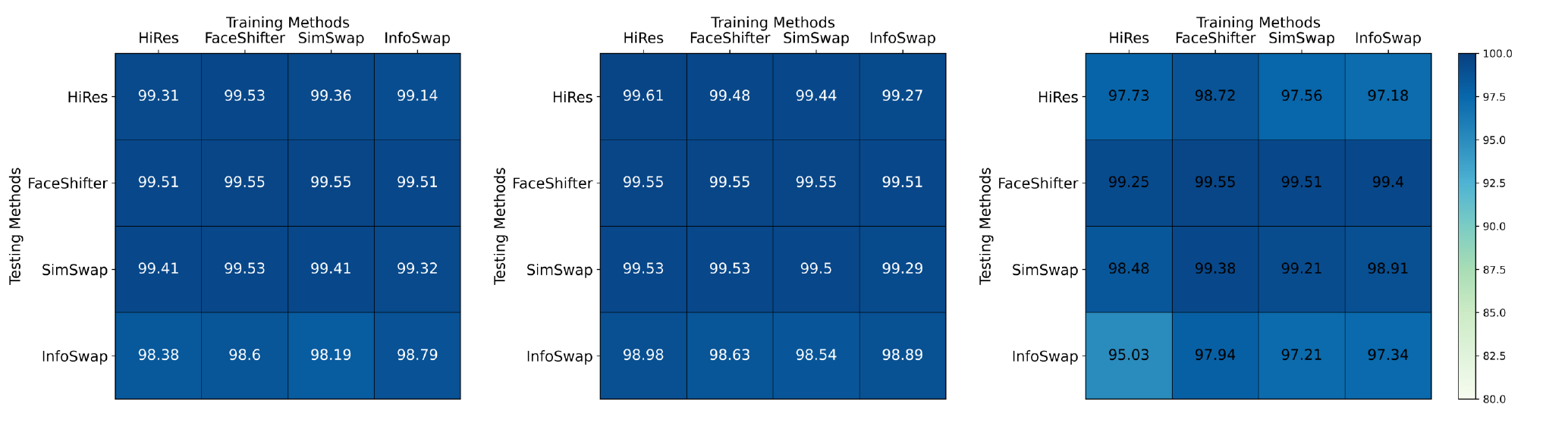}
\vspace{-1.7em}
\caption{Transferability to face-swapping methods that \textbf{explicitly} disentangle identity and attribute information, different figures represent different scenarios. Left: Full-reference scenario (S1); Middle: Half-reference scenario (S2); Right: None-reference scenario (S3). We adopt Top-10 ACC ($\uparrow$) as the metrics. \name still exhibits transferability ability on these methods.}
\label{fig:generalization-1-top10}
\end{figure*}

\begin{figure*}[ht]
\centering
\includegraphics[width=17cm]{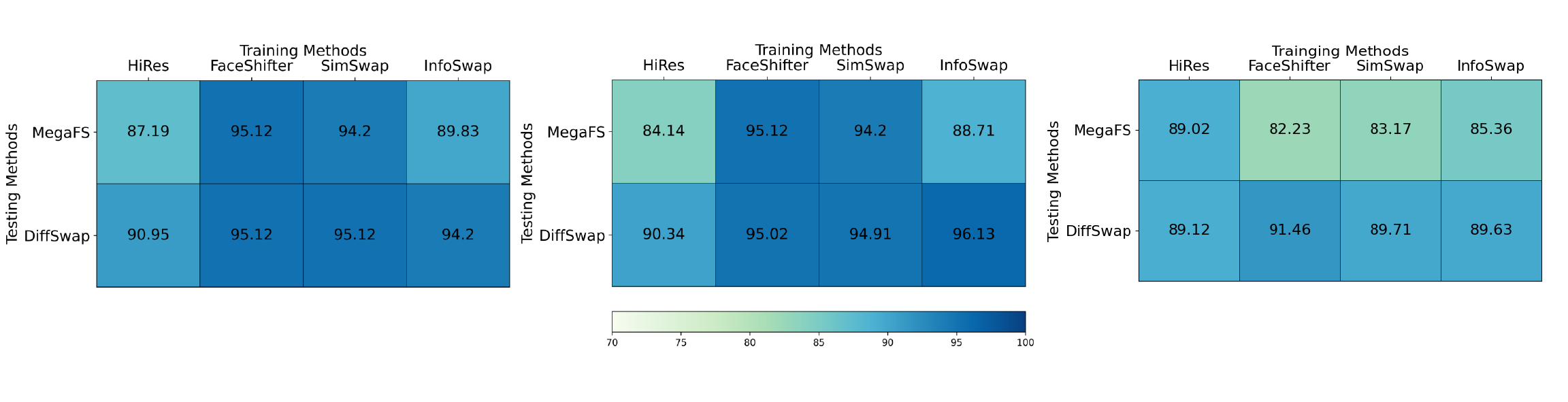}
\vspace{-2em}
\caption{Transferability to face-swapping methods that \textbf{implicitly} disentangle identity and attribute information, different figures represent different scenarios. Left: Full-reference scenario (S1); Middle: Half-reference scenario (S2); Right: None-reference scenario (S3). We adopt Top-10 ACC ($\uparrow$) as the metrics. \name still exhibits transferability ability on these methods.}
\label{fig:generalization-2-top10}
\end{figure*}

For the transferability performance of \name to face-swapping methods that explicitly disentangle identity and attribute information, we further provide Top-5 and Top-10 results in Figure~\ref{fig:generalization-1-top5} and Figure~\ref{fig:generalization-1-top10}, respectively. Similarly, for the transferability performance of \name to face-swapping methods that implicitly disentangle identity and attribute information, we further provide Top-5 and Top-10 results  in Figure~\ref{fig:generalization-2-top5} and Figure~\ref{fig:generalization-2-top10}, respectively.

\subsection{Extension to Swapped Face Videos}
\label{app:ffpp}

\begin{figure*}
\centering
\includegraphics[width=17cm]{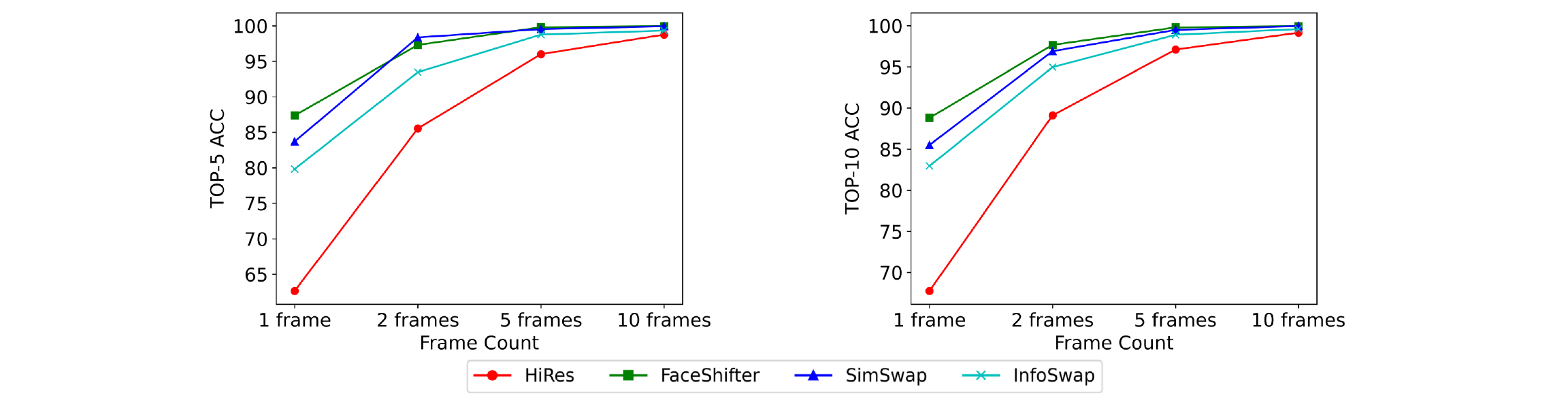}
\caption{Performance of \name when extents to swapped face videos under S1 scenario(refer to Fig.11 of the main paper). Although \name performs not very effective with 1 frame input due to the video quality of FF++, its performance improves significantly as the number of input frames increases. We adopt Top-5 ACC (left) and Top-10 ACC (right) as the metrics.}
\label{fig:ffpp-top5}
\end{figure*}

In Figures~\ref{fig:ffpp-top5}, we illustrated the variations in Top-5 and Top-10 Accuracy of \name trained on different face-swapping methods when testing on the FF++ dataset. 

\subsection{Enhanced Continuous Face-Swapping Attacks}
\label{app:adaptive}

\begin{table}
\caption{Performance of \name against enhanced continuous face swapping.}
\label{tab:adaptive-enhance}
\centering
\begin{tabular}{cccccc}
\hline
Training Method & Metric(\%) & S1 & S2 & S3 \\
\hline
\multirow{3}*{HiRes~\cite{HiRes}} & Top-1 ACC ($\uparrow$) & 95.81 & 96.93 & 78.63 \\
~ & Top-5 ACC ($\uparrow$) & 97.57 & 98.45 & 88.27 \\
~ & Top-10 ACC ($\uparrow$) & 98.10 & 98.73 & 91.09 \\
\hline
\multirow{3}*{FaceShifter~\cite{faceshifter}} & Top-1 ACC ($\uparrow$) & 96.64 & 96.69 & 86.01 \\
~ & Top-5 ACC ($\uparrow$) & 97.95 & 97.95 & 92.15 \\
~ & Top-10 ACC ($\uparrow$) & 98.22 & 98.24 & 93.48 \\
\hline
\multirow{3}*{SimSwap~\cite{simswap}} & Top-1 ACC ($\uparrow$) & 97.12 & 97.02 & 80.73 \\
~ & Top-5 ACC ($\uparrow$) & 98.14 & 98.19 & 88.67 \\
~ & Top-10 ACC ($\uparrow$) & 98.43 & 98.39 & 90.80 \\
\hline
\multirow{3}*{InfoSwap~\cite{infoswap}} & Top-1 ACC ($\uparrow$) & 97.10 & 97.29 & 81.70 \\
~ & Top-5 ACC ($\uparrow$) & 98.12 & 98.26 & 89.91 \\
~ & Top-10 ACC ($\uparrow$) & 98.35 & 98.43 & 92.25 \\
\hline
\end{tabular}
\end{table}

In Section 5.9 of the main paper, we considered the an adversary that employs the same face-swapping method for both iterations of face-swapping. In this section, we further enhance the capabilities of the adversary. We suppose that the adaptive adversary will use two different face-swapping methods, following the approach outlined in the paper, to swap the source face with two different target faces. The two face-swapping methods used were randomly selected from Hires, Faceshifter, SimSwap and InfoSwap. We demonstrate \name's ability to extract the identity information of the source individual in Table~\ref{tab:adaptive-enhance}. It can be observed that even in this scenario, \name still exhibits excellent performance.

\begin{table}[H]
\caption{Parameters of the distortions used in Section 5.5 of the main paper.}
\label{tab:distortion}
\begin{center}
\begin{tabular}{c|c}
\textbf{Distortion} & \textbf{Parameters of Level $k$} \\ \hline
\multirow{2}{*}{Color Jittering} & $b=c=s=0.025k$  \\
& $h=(-0.005k,0.005k)^*$ \\ \hline
Gaussian Noise      & $var=0.002k$                    \\ \hline
Gaussian Blur       & $sigma=\frac{2k-1}{6},kernel=2k-1 $       \\ \hline
Median Blur         & $kernel=2k-1$                       \\ \hline
Salt\&Pepper Noise    & $prob=0.001k$                     \\ \hline
JPEG Compression    & $Q=20k$    \\ \hline
\end{tabular}
\end{center}
\footnotesize{\textsuperscript{*}Here $b$ stands for brightness, $c$ stands for constrast, $s$ stands for saturation and $h$ stands for hue.}
\end{table}

\section{More Details}

\subsection{Distortions}
\label{app:distortion}

Table~\ref{tab:distortion} demonstrates the parameters of the distortions used in Section 5.5 of the main paper. Noted that all distortions except color jittering are applied to images normalized to $(0,1)$.

\ifCLASSOPTIONcaptionsoff
  \newpage
\fi

\end{document}